\documentclass{article} 
\usepackage[preprint]{colm2026_conference}

\usepackage{microtype}
\usepackage{hyperref}
\usepackage{url}
\usepackage{booktabs}
\usepackage{siunitx}

\usepackage{lineno}

\definecolor{darkblue}{rgb}{0, 0, 0.5}
\hypersetup{colorlinks=true, citecolor=darkblue, linkcolor=darkblue, urlcolor=darkblue}

\definecolor{vampire}{HTML}{c41110}

\title{
\textcolor{vampire}{\dataset{}}: Hunting for the Actions Users Want Deep\\Research Agents to Execute
}

\author{Nishant Balepur$^{1, 2, 3*}$, Malachi Hamada$^{3}$, Varsha Kishore$^{3}$, Sergey Feldman$^{3}$,\\ \textbf{Amanpreet Singh}$^{3}$, \textbf{Pao Siangliulue}$^{3}$, \textbf{Joseph Chee Chang}$^{3}$, \textbf{Rachel Rudinger}$^{1}$,\\ \textbf{Eunsol Choi}$^{2}$, \textbf{Jordan Boyd-Graber}$^{1}$, \textbf{Doug Downey}$^{3}$, \textbf{Aakanksha Naik}$^{3}$\\[0.5em]
\small{$^{1}$University of Maryland  
\; $^{2}$New York University \; $^{3}$Allen Institute for Artificial Intelligence (Ai2) }\\[0.5em]
\texttt{\url{nbalepur@umd.edu}, \url{aakankshan@allenai.org}}\\
}

\usepackage[a-1b]{pdfx}

\usepackage{framed}
\usepackage{wrapfig}
\usepackage{mdwlist}
\usepackage{xspace}
\usepackage{siunitx}
\usepackage{latexsym}
\usepackage{colortbl}
\usepackage{xcolor}
\usepackage{nicefrac}
\usepackage{booktabs}
\usepackage{fnpct}
\usepackage{amsfonts}
\usepackage[T1]{fontenc}
\usepackage{bold-extra}
\usepackage{amsmath}
\usepackage{amssymb}
\usepackage{bm}
\usepackage{graphicx}
\usepackage{mathtools}
\usepackage{microtype}
\usepackage{multirow}
\usepackage{multicol}
\usepackage{xpatch}
\usepackage{latexsym,comment}
\usepackage[normalem]{ulem}

\usepackage{booktabs,longtable,array,xcolor, caption}

\usepackage[dvipsnames]{xcolor}

\newcommand*{\missingreference}{{\Huge \colorbox{red}{?reference?}}}
\newcommand*{\missingcitation}{{\Huge \colorbox{red}{?citation?}}}

\makeatletter
\xpatchcmd{\@setref}{\bfseries}{\missingreference}{}{}
\def\@citex[#1]#2{\leavevmode
    \let\@citea\@empty
    \@cite{\@for\@citeb:=#2\do
        {\@citea\def\@citea{,\penalty\@m\ }%
            \edef\@citeb{\expandafter\@firstofone\@citeb\@empty}%
            \if@filesw\immediate\write\@auxout{\string\citation{\@citeb}}\fi
            \@ifundefined{b@\@citeb}{\hbox{\reset@font\missingcitation}%
                \G@refundefinedtrue
                \@latex@warning
                {Citation `\@citeb' on page \thepage \space undefined}}%
            {\@cite@ofmt{\csname b@\@citeb\endcsname}}}}{#1}}
\makeatother

\newcommand{\model}[0]{SQA\xspace}
\newcommand{\modelFull}[0]{ScholarQA\xspace}
\newcommand{\dataset}[0]{DRACULA\xspace}

\newcommand{\dr}[0]{DR\xspace}

\newcommand{\gem}[1]{\mbox{\textsc{gem}}}
\newcommand{\abr}[1]{\textsc{#1}\xspace}

\newcommand{\generic}{\texttt{generic}}
\newcommand{\paper}{\texttt{paper}}

\newcommand{\hidetext}[1]{}
\newcommand{\ignore}[1]{}

\usepackage{cleveref}
\usepackage{microtype}
\usepackage[most]{tcolorbox}

\usepackage{enumitem}
\crefformat{section}{\S#2#1#3}
\crefformat{subsection}{\S#2#1#3}
\crefformat{subsubsection}{\S#2#1#3}

\newif\ifcomment\commenttrue

\ifcomment
    \newcommand{\pinaforecomment}[3]{\colorbox{#1}{\parbox{.8\linewidth}{#2: #3}}}

    \newcommand{\prtodo}[1]{\pinaforecomment{lightblue}{pr}{#1}}
    \newcommand{\prtodoi}[1]{\pinaforecomment{lightblue}{pr}{#1}}
\else
    \newcommand{\pinaforecomment}[3]{}
    \newcommand{\prtodo}[1]{}
    \newcommand{\prtodoi}[1]{}
\fi

\newcommand{\smallurl}[1]{ \begin{tiny}\url{#1}\end{tiny}}

\definecolor{lightblue}{HTML}{3cc7ea}
\definecolor{CUgold}{HTML}{CFB87C}
\definecolor{grey}{rgb}{0.95,0.95,0.95}
\definecolor{ceil}{rgb}{0.57, 0.63, 0.81}
\definecolor{UMDred}{HTML}{ed1c24}
\definecolor{UMDyellow}{HTML}{ffc20e}

\definecolor{SkyBlue}{rgb}{0.53, 0.81, 0.92}

\newtcolorbox[
  list inside=prompt,
  auto counter,
  number within=section
]{prompt}[1][]{%
  enhanced,
  float*=t, 
  colbacktitle=black!60,
  fonttitle=\small,
  coltitle=white,
  fontupper=\footnotesize,
  boxsep=4pt,
  left=0pt, right=0pt, top=0pt, bottom=0pt,
  boxrule=1pt,
  width=\textwidth,          
  enlarge left by=0mm,
  enlarge right by=0mm,
  listing only,
  listing options={
    basicstyle=\ttfamily\footnotesize,
    breaklines=true,
    breakatwhitespace=true,
    language=json
  },
  #1,
}

\newtcolorbox[
  list inside=trace,
  auto counter,
  number within=section
]{trace}[1][]{%
  enhanced,
  float*=t,
  colback=blue!5,             
  colbacktitle=blue!60!black, 
  colframe=blue!60!black,     
  fonttitle=\small,
  coltitle=white,
  fontupper=\footnotesize,
  boxsep=4pt,
  left=0pt, right=0pt, top=0pt, bottom=0pt,
  boxrule=1pt,
  width=\textwidth,
  enlarge left by=0mm,
  enlarge right by=0mm,
  listing only,
  listing options={
    basicstyle=\ttfamily\footnotesize,
    breaklines=true,
    breakatwhitespace=true,
    language=json
  },
  #1,
}

\begin{document}

\ifcolmsubmission
\linenumbers
\fi

\maketitle

\begin{abstract}


Scientific Deep Research (\dr{}) agents answer user queries by synthesizing~research papers into multi-section reports.
User feedback can improve their utility, but~existing protocols only score the final report, making it hard to study and learn~which~intermediate actions \dr{} agents should take to improve reports.
We collect \dataset{}, the~first dataset with user feedback on intermediate actions for \dr{}.
Over five weeks, nineteen expert CS~researchers ask queries to a \dr{} system that proposes actions (e.g.,~``\textit{Add a section on datasets}'').
Our users select actions they prefer, then judge whether an output report applied their selections successfully, yielding 8,103 action preferences and 5,230 execution judgments. 
After confirming a \dr{} agent can execute \dataset{}'s actions, we study the predictability of user-preferred actions via simulation---how well LLMs predict the actions users select---a step toward~learning to generate useful~actions.
We discover:
(1) LLM judges initially~struggle to predict action selections, but improve most when using a user's full selection~history, rather than self-reported or extrapolated user context signals;
(2) Users' selections for the same query differ based on unstated goals, bottlenecking simulation and motivating affordances that let users steer reports;
and (3) Our simulation results inform an online intervention that generates new actions based on the user's past interactions, which users pick most often in follow-up studies.
Overall, while work extensively studies execution, \dataset{} reveals a key challenge is deciding which actions to execute in the first place.
We~open-source~\dataset{}'s study design, user feedback, and simulation~tasks to spur work on action feedback~for~long-horizon~agents.\footnote{Our data and code are available at: \url{https://github.com/allenai/dracula}.}

\begingroup
\renewcommand{\thefootnote}{\fnsymbol{footnote}}
\footnotetext[1]{Work primarily completed during internship at Ai2.}
\endgroup



\end{abstract}


\section{Introduction: User Feedback in Deep Research Could Use More \textcolor{vampire}{Bite}}

Scientific Deep Research (\dr{}) agents synthesize scientific papers into long-form reports in response to user queries \citep{Asai2024OpenScholarSS}.
Even when \dr{} agents excel~on offline~benchmarks \citep{wei2025browsecomp}, user feedback remains key to improving their utility \citep{ouyang2022training}.

Most work collects user feedback on the final \dr{} report with pairwise \citep{zhao2025sciarena} or~absolute ratings \citep{hwang2026deep}.
However, such approaches do not let users directly rate the usefulness of intermediate actions taken during generation---like which papers to retrieve, which sections to add, or how to explain ideas---obscuring the actions users prefer.
Thus, we lack datasets to analyze and learn which intermediate actions are useful for \dr{}.



\begin{figure}[t]
    \centering
    \includegraphics[width=\linewidth]{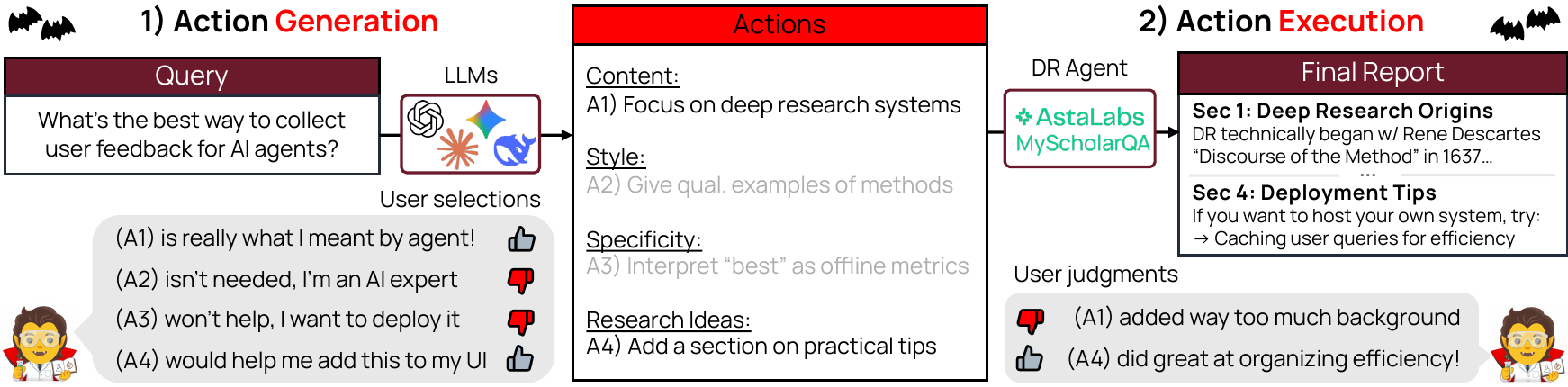}
    \vspace{-3ex}
    \caption{\small \textbf{Overview of \dataset{}'s feedback.} Rather than only judging reports, we reveal~actions users want \dr{} to take via two steps: (1) LLMs propose actions that modify reports for users to pick from; and (2) the MyScholarQA DR agent writes a report following selected actions that users judge for execution quality. Over 450 hours, we curate 13,333 query–action-judgment pairs from 19 researchers.}
    \label{fig:main}
    \vspace{-1.5ex}
\end{figure}

We release \textbf{\dataset{}}:\footnote{To spare readers our backronym: \textbf{D}eep \textbf{R}esearch \textbf{A}ction \textbf{C}orpus via \textbf{U}ser-in-the-\textbf{L}oop \textbf{A}nnotation} the first open-source dataset with user feedback for understanding the actions users want \dr{} systems to take.~Over five weeks and 450 annotation hours,~19 researchers ask queries to a \dr{} system \citep{singh-etal-2025-ai2} that proposes \textbf{actions}---decisions the agent can take to improve how reports answer user queries (e.g., ``\textit{Explain concepts with analogies}'', ``\textit{Add a section on datasets}'').
For each query, users pick the actions they~want~for their report, then view a report tailored to their selections to judge whether actions~were executed effectively (Figure~\ref{fig:main}).
In total, \dataset{} has 8,103 binary selections as preferences on proposed actions and 5,230 binary judgments for action execution, each with~rationales~(\cref{section:dataset}).

We first ensure \dr{} systems can realistically execute the actions in \dataset{}.
Users rate our tested models as more successful at executing selected actions than at proposing useful ones (\cref{subsection:data_summary}), and qualitative analyses reveal system limitations are a rare reason for rejecting actions (\cref{subsection:coding}).
Thus, a key challenge in \dr{} is not just executing actions, but knowing \textit{which} actions users want executed---motivating the rest of our paper's focus on studying the latter.






To further analyze user action selections, we measure their predictability via simulation~experiments \citep{seshadri2026lost}: given an action, can LLMs predict whether users select it, and what improves these predictions?
Practically, reliable predictions can serve as training rewards \citep{shao2025dr} and benchmark metrics \citep{liang2025towards} to improve \dr{} actions.

Baseline few-shot prompting with five frontier LLM judges initially struggles to predict the actions users select (\cref{subsection:initial_predict}).
Deeper experiments show predictions~improve most when using the user's entire past selection history rather than extrapolated, self-reported, and cold-start user contexts like research papers and stated preferences---indicating simulations benefit most from user context closer to the task format (\cref{subsection:follow_up_predict}).
To determine whether our models have reached simulation's upper bound, we run a follow-up study and show~users disagree with their past selections as their goals change (\cref{subsection:unstable}).
Thus, certain users' preferences are inherently unpredictable and may always need affordances to clarify them in \dr{} systems.

Finally, motivated by the benefits of user-specific context in action prediction, we evaluate whether three sources of user context---past research papers, stated preferences, and revealed preferences from past selections---can improve action generation.
Users most often select actions generated from their revealed preferences (\cref{subsection:user_context}), highlighting the value of learning from past interactions to improve \dr{} actions, and more generally, the potential for offline simulation experiments to inform online interventions that can make agents more useful.


Overall, while researchers extensively study how well \dr{} systems can execute actions, we show a key bottleneck is predicting which actions users~want in the first place---a difficult task that largely benefits from user-specific modeling.
More broadly, as execution horizons scale, agents must take increasingly complex actions to support users, making intermediate feedback beyond the final output key for improving their utility. Our contributions~are:

\begin{enumerate}[leftmargin=1.5em, itemsep=0em, topsep=0em]
    \item \dataset{}, the first large-scale dataset of user feedback on intermediate \dr{} actions, including queries, actions, user selections, execution judgments, and user contexts.
    \item A new simulation task for LLMs to predict the actions users want \dr{} systems to take.
    \item Evidence that selected actions are highly-user specific and change based on user~goals, and that simulation experiments can guide online interventions to make \dr{} more~useful.
\end{enumerate}


\section{\dataset{}: \textcolor{vampire}{Inviting} Users for Feedback on Deep Research Actions} \label{section:dataset}

To improve \dr{}, we rely on user feedback that can teach systems how to generate reports users prefer \citep{ouyang2022training}.
Most \dr{} feedback only scores the report \citep{zhao2025sciarena}, but producing reports requires many intermediate actions, like which papers to~retrieve, which sections to include, and how to explain ideas \citep{pan2026interdeepresearch}.
Report-level feedback alone provides little insight on which of these actions users prefer \citep{movva2026whats}, making it difficult to study and learn which actions \dr{} should take to support users.

To address this, we explicitly gather feedback on natural language actions that describe~how \dr{} agents could improve reports (e.g., ``\textit{Find new datasets}'', ``\textit{Add deployment tips}'').~In~our feedback protocol, users issue a query and we propose actions for them to select from.
We then show a report reflecting the selected actions, which users judge for execution quality (Figure~\ref{fig:main}).
We host our annotation in a \dr{} interface to build \dataset{}: a dataset of \dr{} queries, actions, and feedback split into action selections and executions, described next.


\textbf{Query Selection:}
To reflect real \dr{} usage without manually writing queries,~annotators select 50 queries of interest from a set asked by users in the deployed \textsc{ScholarQA} system \citep{singh-etal-2025-ai2}.
We label all queries by field/subfield with Gemini-2.5 Flash (e.g., ``AI''/``AI Agents'') to help annotators search.
For each query $q$, annotators self-report their intent for selecting $q$---paper search, learning, brainstorming, implementation design, or writing---and expertise from 1--5.


\textbf{Action Generation:}
For each $q$, we randomly prompt one of GPT-4.1, Claude-4 Sonnet, Gemini-2.5 Flash, or DeepSeek-V3 to create eight actions $\mathcal{A}=\{a_1,\dots,a_8\}$ (Prompt~\ref{prompt:action_generation_generic}).
We use actions as they clearly describe how a \dr{} system will attempt to improve a report,~unlike other designs such as follow-up questions that focus on clarifying intent \citep{wang2025liveresearchbench}.


Prior work shows user context improves clarifications \citep{feng2023towards, montazeralghaem2025asking}, so we test whether this benefits action generation.
We use research papers~as user~context, commonly studied in scientific tasks similar to \dr{} \citep{Kreutz2022ScientificPRA, Mysore2023EditableUPA}.  
For each $q$, we create four \textbf{\generic} actions that condition on $q$, and~four \textbf{\paper} actions that also condition on annotator-picked papers $\mathcal{P}$ reflecting their interests.~Varied prompt contexts also boosts output diversity \citep{kim-etal-2023-concept}, desirable in feedback data \citep{lambert2024tulu}.
We de-duplicate actions across conditions with GPT-4.1 (Prompt~\ref{prompt:deduplication}).

Our actions span four qualitative categories (the center of Figure~\ref{fig:main} has examples) with~common instruction types that users add to queries to improve \dr{} reports~\citep{haddad2026understanding}: \textit{Content} (what the report covers), \textit{Style} (how the report is written), \textit{Specificity} (how to treat the query scope), and \textit{Research Ideas} (how to adapt to the researcher's goals).
To ensure~coverage, we generate one \generic{} and one \paper{} action per category (i.e., four each, eight in total).


Annotators view $\mathcal{A}$ as a checklist (Figure~\ref{appendix:fig:plan}) grouped by category and mark which actions $\mathcal{A}' \subseteq \mathcal{A}$ they want the agent to take and a rationale, yielding $l_{\text{select}} \in \{0, 1\}$ (Figure~\ref{fig:main}, left).



\textbf{Report Generation:}
We use the \textsc{MyScholarQA} system from \citet{balepur2026dont}, backed by Claude-4 Sonnet.\footnote{The system is open-sourced with a demo: \url{https://personalized-scholarqa.apps.allenai.org/}.}
It creates a report to answer $q$ by retrieving Semantic Scholar papers, re-ranking, extracting evidence, planning outlines, and generating sections.
MyScholarQA~alters \modelFull{} to use $\mathcal{A}'$ as input at each step (Appendix~\ref{appendix:section:model}).
To aid annotation, we highlight spans in the report where each action is applied (Figure~\ref{appendix:fig:report}).
Annotators judge whether each $a \in \mathcal{A}'$ was executed well and a rationale, giving $l_{\text{exec}} \in \{0, 1\}$ (Figure~\ref{fig:main},~right).


\textbf{User Study:}
We recruit 20 CS researchers in diverse fields (e.g., Security, CV, NLP, CompBio, AI Ethics) from Upwork, verified as active \dr{} users via a pilot survey. 
We run a~five week study over \textasciitilde450 hours at \$30--45/hour, approved by~our~IRB.
As quality control, we omit ten queries that are not in scope for DR (e.g., ``\textit{How to DR}''), and one annotator who reused rationales, yielding 8,103 selections for $l_{\text{select}}$ and 5,230 judgments for $l_{\text{exec}}$.\footnote{In total, \dataset{} contains 902 unique queries and 7731 unique actions.}


\section{Generating Actions is a High-\textcolor{vampire}{Stakes} Challenge for Deep Research} \label{section:generation}

We first verify that \dr{} systems can execute the actions in \dataset{}, ensuring execution quality does not largely impact our future analyses of user preferences on actions.
We first compare how well models generate versus execute actions, showing that they struggle more on the former (\cref{subsection:data_summary}).
Qualitative analyses further reveal that system limitations are a rare reason for rejecting actions (\cref{subsection:coding}), motivating our later studies on user-preferred actions~(\cref{section:experiments}).

\subsection{Neglecting Action Generation Can \textcolor{vampire}{Drain} the Usefulness of Deep Research} \label{subsection:data_summary}

\begin{wrapfigure}{r}{0.42\linewidth}
    \centering
    \vspace{-1ex}
    \includegraphics[width=\linewidth]{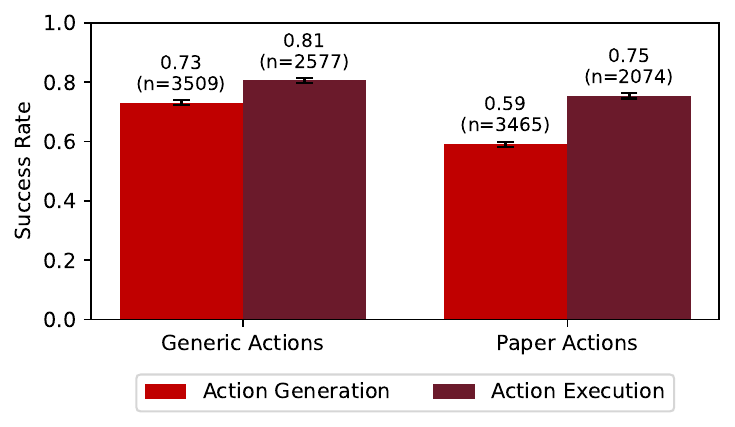}
    \vspace{-4.5ex}
    \caption{\small Action generation and execution success over action types. Models cannot infer all the \dr{} decisions users find useful.}
    \label{fig:action_gen_vs_exec}
    \vspace{-1ex}
\end{wrapfigure}

The proportion of actions that users rate as well-executed in reports exceeds the~proportion of actions they pick (Figure~\ref{fig:action_gen_vs_exec}): LLMs struggle to infer actions users~want,~but more reliably execute actions once specified.
Extensive work aims to improve \dr{} executions~\citep{wei2025browsecomp}, but another key challenge lies in learning what to execute---a gap \dataset{} fills.

We also note despite prior work showing the benefits of using papers as user context \citep{Mysore2023EditableUPA}, \paper{} actions are selected less than~\generic{}.
\cref{subsection:user_context} tests this further by studying more user contexts.



\begin{figure}[h!]
    \centering
    \includegraphics[width=\linewidth]{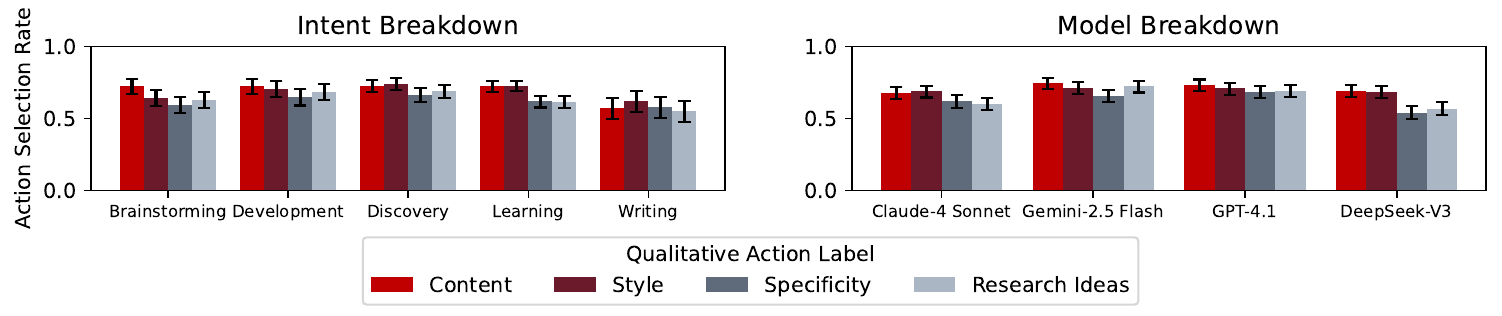}
    \vspace{-4ex}
    \caption{\small Breakdown of action selection rates across query intent, model, and qualitative type. Actions on content and style are selected most often, and LLMs struggle to produce actions for writing intents.}
    \label{fig:breakdown}
\end{figure}

As initial analyses, we group action selection~rates by user intent, action type, and generator LLM.
Actions altering report content and style are generally preferred over specificity or research ideas; Claude and DeepSeek generate the latter less effectively (Figure~\ref{fig:breakdown}, right).
Writing intent queries have lower selection rates, but intuitively, users prefer style actions more often (Figure~\ref{fig:breakdown}, left).
Notably, Claude-4 Sonnet backs report generation in MyScholarQA but this LLM still has low action selection rates, so action generation being less successful than execution holds when controlling for LLM.


\subsection{Qualitative Analysis: An Initial \textcolor{vampire}{Taste} of Action Generation} \label{subsection:coding}

To study and better illustrate the data in \dataset{}, we run qualitative analyses under~two research~questions: RQ1---\textit{Which actions do users select?} and RQ2---\textit{Why are actions not selected?}

\textbf{RQ1---What actions do users select?} We run a semi-automated qualitative coding procedure \citep{bingham2023data, lam2024concept}.
Claude-4.6 Opus infers clusters with high-level action types from 200 random actions per user, then merges duplicate clusters.
Two authors refine clusters, then Gemini-3 Flash assigns each action to the most relevant cluster (Appendix~\ref{appendix:section:clustering}).

The most often selected actions summarize key points and compare papers (Table~\ref{table:qualitative_coding_actions}, top), aiding comprehension \citep{erera-etal-2019-summarization}, while the least selected propose practical tools and cross-domain connections (Table~\ref{table:qualitative_coding_actions}, bottom).
From users' rationales, tools were often rejected due to intent misalignment (e.g., ``\textit{No code snippets are required, as this is a theoretical question}'').
Cross-domain connections were rejected when models over-personalize to user papers, degrading clarity (e.g., ``\textit{This is related to my research but I don't think what the model is suggesting is even a thing}'').
In only 19 cases, users cited \dr{} system limitations as the reason for rejecting an action and never showed ``miswanting'' \citep{gilbert2000miswanting}---realizing they did not want an action---suggesting most rejections reflect true action dispreferences.



\textbf{RQ2: Why are actions not selected?} Applying the same coding procedure on rationales for unselected actions, we surface 15 failure modes. 
The most common involve miscalibrating query scope (Table~\ref{table:qualitative_coding_rationales}): actions are rejected when they aggressively narrow search, expand beyond the user's needs, drift off-topic, or neglect user goals.
Thus, action generation relies on user-specific context, which we keep in mind when designing future analyses (\cref{section:experiments}, \cref{subsection:user_context}).

\begin{table}[t]
\scriptsize
\centering
\captionsetup{singlelinecheck=false}
\renewcommand{\arraystretch}{1.0}
\setlength{\tabcolsep}{4pt}

\begin{tabular}{
@{}
>{\raggedright\arraybackslash}p{0.53\textwidth}
>{\raggedright\arraybackslash}p{0.3\textwidth}
>{\centering\arraybackslash}p{0.12\textwidth}
@{}
}

\toprule
\textbf{Action Cluster (Title + Description)} & \textbf{Examples} & \textbf{Selection Rate} \\
\midrule

\textit{Summarize Key Points (N=332).}
Adding summaries, bullet-point takeaways, and executive overviews to help understand the report.
&
Add bulleted takeaways to sections,
Explain unique features of each YOLO
& 0.798 \\

\addlinespace[2pt]

\textit{Comparison Analysis (N=461).}
Adding structured tables and side by side evaluations to help assess strengths, differences, and trade-offs.
&
Compare Tensorflow to PyTorch,
Contrast with Monte Carlo Methods
& 0.774 \\

\midrule
\noalign{\vskip -0.6ex}
\multicolumn{3}{c}{...}\\
\noalign{\vskip -0.9ex}
\midrule

\textit{Add Tools And Resources (N=128).}
Finding specific tools, artifacts, and repositories to support reproducibility and hands-on exploration.
&
Include code resources from papers,
Display public materials databates
& 0.586 \\

\addlinespace[2pt]

\textit{Cross Domain Connections (N=105).}
Connecting the query to other fields to widen context and suggest inter-disciplinary opportunities.
&
Link to Tree of Thoughts integration,
Connect to inverse design workflows & 0.486
\\

\bottomrule
\end{tabular}

\vspace{-1.5ex}
\caption{\small The two most and least often selected action types uncovered via clustering. Users tend to select actions that improve comprehension, but reject ones proposing practical tools that are misaligned with their intent, or over-personalize by suggesting cross-domain connections. The full list is in Table~\ref{appendix:table:action_cluster_full}.} \label{table:qualitative_coding_actions}
\vspace{-3ex}

\normalsize
\end{table}
\begin{table}[t]
\scriptsize
\centering
\captionsetup{singlelinecheck=false}
\renewcommand{\arraystretch}{1.05}
\setlength{\tabcolsep}{4pt}

\begin{tabular}{
@{}
>{\raggedright\arraybackslash}p{0.55\textwidth}
>{\raggedright\arraybackslash}p{0.43\textwidth}
@{}
}

\toprule
\textbf{Rationale Cluster (Title + Description)} & \textbf{Example} \\
\midrule

\textit{Scope Too Narrow (N=397).}
The action restricts the scope more than the user wants, excluding topics and material they consider relevant.
&
Focus on multi-model empirical validation for domain tasks (``\textit{I would like to keep the search broad}'') \\

\addlinespace[2pt]

\textit{Topic Not Relevant (N=326).}
The action targets a topic or domain unrelated to the query and does not contribute to their objective.
&
List open LLM-dynamics questions. (``\textit{My question doesn't have much to do with dynamical systems.}'') \\

\addlinespace[2pt]

\textit{Misaligned With User Goals (N=294).}
The action is in the right domain but focuses on the wrong aspect relative to what the user truly wants.
&
Benchmark across video datasets and domains (``\textit{No need for benchmarks, I want to see how to implement it}'') \\

\addlinespace[2pt]

\textit{Exceeds Needed Scope (N=261).}
The action expands the scope beyond what is necessary, adding tangential topics or follow-up directions.
&
Find papers using logit lens with other interpretability techniques (``\textit{No need, I am just curious about logit lens}'') \\

\bottomrule
\end{tabular}

\vspace{-1.5ex}
\caption{\small The four most common rationale categories that explain why users rejected actions uncovered via clustering. Most failures relate to miscalibrating query scope. The full lists are in Tables~\ref{appendix:table:bad_rationale_cluster_full} and~\ref{appendix:table:good_rationale_cluster_full}.}
\label{table:qualitative_coding_rationales}
\vspace{-1ex}

\normalsize
\end{table}



\section{\textcolor{vampire}{Mirroring} User Feedback on Action Selection} \label{section:experiments}




We analyze user preferences on \dataset{}'s actions via user simulations: given an action, can LLMs predict whether users will select it?
Such analyses reveal how predictable user preferences are,~and what drives this predictability.
Practically, user simulations are more scalable than real user feedback \citep{wu2025collabllm}, so reliable predictions could form training rewards \citep{shao2025dr} and benchmark metrics \citep{liang2025towards} to improve \dr{} actions.

We turn \dataset{} into reusable evaluations that test whether LLMs can predict the actions users select (\cref{subsection:setup}).
We first run prompt baselines, showing~predicting action~selection initially struggles (\cref{subsection:initial_predict}).
We then test varied modeling choices to improve predictions~and reveal they improve most when using a user's past selections (\cref{subsection:follow_up_predict}).
We finally see whether~models have reached the inherent predictability in \dataset{} by studying user consistency (\cref{subsection:unstable}).


\subsection{\textcolor{vampire}{Turning} \dataset{} into a User Simulation Task} \label{subsection:setup}

We define a classification task called \textit{action prediction} for simulating user selections: Given a query $q$ and proposed action $a$, predict $\hat{l}_{\text{select}} \in \{0,1\}$, indicating whether the user selected~$a$.


We sort our data chronologically (oldest to newest) and use 80/10/10 splits based on~$q$, giving 5982/1140/918 examples for train/dev/test~with a 0.32/0.68 test split for 0/1 labels.
We omit noisy test set items where GPT-5 cannot~predict the true label via the user's rationale or where we flag low-quality queries (\cref{section:dataset}); this impacts <5\% of the data.
We report Macro-F1 score---the unweighted average of per-class F1---commonly used for binary judges in \dr{} \citep{sharma2025researchrubrics}, and standard error via bootstrap sampling with $n=500$ \citep{efron1992bootstrap}.


\begin{figure}[h]
    \centering
    \includegraphics[width=\linewidth]{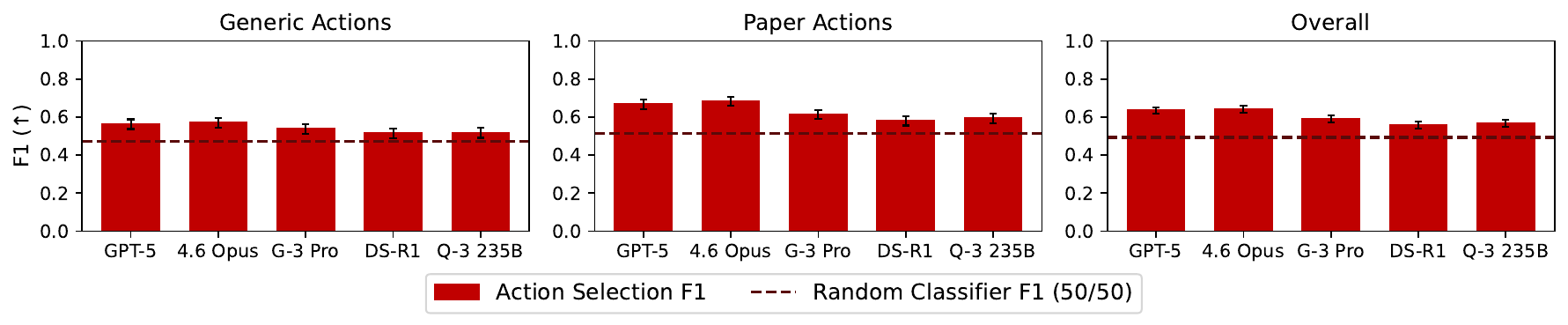}
    \vspace{-4ex}
    \caption{\small Macro F1 score of five few-shot LLM judges on action prediction and a random binary classifier (50/50).
    F1 does not largely exceed random on \generic{} actions, showing room for improvement.}
    \label{fig:prediction}
    \vspace{-1ex}
\end{figure}

\subsection{Evaluating Action Prediction Off the \textcolor{vampire}{Bat}} \label{subsection:initial_predict}

We assess five strong baseline LLM judges on action prediction: GPT-5, Claude-4.6 Opus, Gemini-3.0 Pro, DeepSeek-R1, and Qwen-3 235B Thinking (Appendix~\ref{appendix:section:model}).
The prompt describes the task and requests a JSON with a 0/1 prediction and explanation (Prompts~\ref{prompt:action_prediction}), and uses 1000 few-shot $(q,a,l_{\text{select}})$ training examples randomly sampled across~all users.

F1 scores do not largely exceed a random 50/50 baseline across LLMs (Figure~\ref{fig:prediction}), especially on \generic{} actions, showing judges have room to improve in~action prediction; we explore this next (\cref{subsection:follow_up_predict}). 
Action prediction on \paper{} actions has higher F1, so LLMs may more~reliably distinguish differences in preferred versus dispreferred actions that tailor to user context.


\subsection{\textcolor{vampire}{Awakening} Action Prediction Capabilities} \label{subsection:follow_up_predict}


To improve action prediction (\cref{subsection:initial_predict}), we now assess different modeling choices.
Motivated~by user-specific action generation flaws in \cref{subsection:coding}, we test adding user context, along with general analyses on trivial baselines, retrieval, and task engineering.
To limit costs, we use~GPT-5 only, a strong LLM in Figure~\ref{fig:prediction}, but Appendix~\ref{appendix:section:compare_prediction} runs a subset of analyses with more LLMs.

\textbf{Action Prediction is Non-Trivial.}
We verify the task is unsolvable via simple cues.
Trivial baselines---GPT-5 prompted without the query and action as partial-input baselines \citep{poliak-etal-2018-hypothesis}, and an SVM trained with TF-IDF embeddings \citep{sparck1972statistical}---score near random (Table~\ref{table:ablations}, \textit{Trivial Baselines}), so action prediction needs more than shallow reasoning.

\textbf{User History is Key.} As action selection varies by user (\cref{subsection:data_summary}), we test whether conditioning on past selections from just the target user aids predictions.
This largely boosts F1~(Table~\ref{table:ablations}, \textit{User-Specific History} vs \textit{All User History}), so user modeling benefits the task.
Fewer examples and simple user selection priors do not explain these gains: the same number of examples over all user selections or just giving selection rates (e.g., ``this user picks 71\% of actions'') lower F1 (Table~\ref{table:ablations}, \textit{Context Engineering}), so having the text of actions is useful ~\citep{shaikh2026learning}.
We thus adopt user-specific history as the baseline for the remaining experiments.

\setlength{\fboxsep}{1pt}

\newcommand{\err}[2]{%
  \ensuremath{%
    \text{#1}\,\text{\scriptsize$\pm\,\num[round-mode=figures,round-precision=1]{#2}$}%
  }%
}

\begin{table}[t]
\small
\centering
\setlength{\tabcolsep}{4.5pt}
\begin{tabular}{@{}rlccc@{}}
\toprule
\multicolumn{1}{l}{} Experiment Type &
  Configuration &
  \multicolumn{1}{l}{F1 (Generic)} &
  \multicolumn{1}{l}{F1 (Paper)} &
  \multicolumn{1}{l}{F1 ($\text{Avg}_\text{macro}$)} \\ \midrule
  
\textit{Initial Attempt}                                                                       & \cellcolor{red!20}All User History                 & \cellcolor{red!20} \err{0.562}{0.026}  & \cellcolor{red!20} \err{0.667}{0.025} & \cellcolor{red!20} \err{0.614}{0.017} \\ \cmidrule(l){2-5} 
\multirow{3}{*}{\textit{\begin{tabular}[c]{@{}r@{}}Trivial\\ Baselines\end{tabular}}}   & No Query                         & \err{0.477}{0.021} & \err{0.527}{0.026} & \err{0.502}{0.017} \\
                                                                                        & No Action                        & \err{0.453}{0.019} & \err{0.435}{0.020} & \err{0.444}{0.015} \\
                                                                                        & SVM + TFIDF                      & \err{0.519}{0.023}  & \err{0.583}{0.026} & \err{0.551}{0.017} \\ \midrule 
\multirow{3}{*}{\textit{\begin{tabular}[c]{@{}r@{}}Context\\ Engineering\end{tabular}}} &
  \cellcolor{red!20}User-Specific History &
  \cellcolor{red!20} \err{0.640}{0.024} &
  \cellcolor{red!20} \err{0.726}{0.023} &
  \cellcolor{red!20} \err{0.683}{0.017} \\
                                                                                        & All User History, Equal Examples & \err{0.572}{0.026}  & \err{0.628}{0.024} & \err{0.600}{0.017} \\
                                                                                        & User Selection Statistics        & \err{0.529}{0.025}  & \err{0.656}{0.024} & \err{0.592}{0.017} \\ \cmidrule(l){2-5} 
\multirow{4}{*}{\textit{\begin{tabular}[c]{@{}r@{}}Feature\\ Engineering\end{tabular}}} & + Papers                        & \err{0.593}{0.026}  & \err{0.571}{0.026} & \err{0.582}{0.018} \\
                                                                                        & + Qualitative Action Labels                & \err{0.619}{0.027}  & \err{0.684}{0.024} & \err{0.651}{0.018} \\
                                                                                        & + Rationales                     & \err{0.634}{0.026}  & \textbf{\err{\textbf{0.733}}{0.024}} & \err{0.684}{0.017} \\
                                                                                        & + Expertise and Intent               & \err{0.615}{0.025}  & \err{0.716}{0.024} & \err{0.665}{0.017} \\ \cmidrule(l){2-5} 
\multirow{2}{*}{\textit{\begin{tabular}[c]{@{}r@{}}Stated\\ Preferences\end{tabular}}}
                                                                                        & User-Stated Rules         & \err{0.586}{0.025}  & \err{0.628}{0.023} & \err{0.603}{0.017} \\
                                                                                        & Model-Inferred Rules             & \err{0.585}{0.025}  & \err{0.676}{0.023} & \err{0.631}{0.017} \\ \cmidrule(l){2-5}

\multirow{3}{*}{\textit{\begin{tabular}[c]{@{}r@{}}Long-Context\\ Management\end{tabular}}} &
  Random Sample ($k=50$) &
  \err{0.620}{0.024} &
  \err{0.662}{0.024} &
  \err{0.641}{0.017} \\
                                                                                        & GritLM Retriever ($k=50$)        & \err{0.640}{0.023}  & \err{0.711}{0.022} & \err{0.675}{0.016} \\
                                                                                        & GPT-5 Retriever ($k=50$)         & \err{0.622}{0.023}  & \err{0.706}{0.022} & \err{0.664}{0.016} \\ \cmidrule(l){2-5} 
\multirow{2}{*}{\textit{\begin{tabular}[c]{@{}r@{}}Task\\ Engineering\end{tabular}}}    & Multi-Classification Task                     & \err{\textbf{0.671}}{0.026}  & \err{0.709}{0.023} & \err{\textbf{0.690}}{0.017} \\
                                                                                        & Pairwise Comparison Task                  & \err{0.457}{0.034}  & \err{0.631}{0.025} & \err{0.544}{0.022} \\ \midrule
                                                                                        \textit{Random Baseline}                                                               & Random Choice (50/50)             & \err{0.476}{0.024}  & \err{0.518}{0.025} & \err{0.497}{0.017} \\ \bottomrule
\end{tabular}
\vspace{-1.5ex}
\caption{\small Analysis of trivial baselines, context engineering, feature engineering, stated preferences, long-context management, and task engineering to boost action prediction with GPT-5. User-specific history largely boosts predictions. \colorbox{red!20}{Rows in red} are baselines for the two sections. Best scores are \textbf{bold}.}
\label{table:ablations}
\vspace{-1.5ex}
\end{table}

\textbf{Other User Context Types Don't Cut It.} We evaluate whether more user context improves action prediction.
Augmenting user-specific history prompts with user-selected papers in~\cref{section:dataset}, action labels (e.g., ``Content'', ``Style''), past rationales, and reported intent/expertise never largely beat the~user history baseline (Table~\ref{table:ablations}, \textit{Feature Engineering}), except for a small but insignificant boost from rationales.
Thus, user behavior that is closer to the task is a~stronger predictive signal for future behavior versus self-reported user descriptors \citep{Binz2024CentaurAF}.

\textbf{Actions Speak Louder than Words.} LLMs may struggle to infer the latent preferences from user's action history that drive their future selections \citep{samuelson1938note}, degrading action predictions.
To test this, we first elicit our users' \texttt{stated preferences}~\citep{hensher1994stated}---five high-level rules our users write to describe their ideal \dr{} system after having interacted with our system (e.g., ``\textit{Introductions must follow academic paper structures}'').
We then use these as input for action prediction and compare this to giving five rules GPT infers from each user's~past selections.
GPT rules are more predictive than user-written rules, but neither exceeds user-specific history (Table~\ref{table:ablations}, \textit{Stated Preferences}).
Thus, capturing how users actually select actions is more predictive than the actions they say they prefer \citep{chung2025literarytaste}.


\textbf{Long Context is Unlikely to be the Main Culprit.} Action prediction models reason over lengthy user selections, raising concerns on long-context limitations \citep{laban2026llms}.
We test this by shortening input context via: 1) randomly sampling 50 examples; or 2) retrieving the top-50 relevant examples via GritLM \citep{muennighoff2025generative} and a GPT-5 reranker \citep{sun-etal-2023-chatgpt}.
Neither method improves F1 (Table~\ref{table:ablations}, \textit{Long-Context Management}), but F1 is not much worse, so selection histories could be compressed while maintaining predictive power for efficiency \citep{sun-etal-2025-persona}. 
Appendix~\ref{appendix:section:retrieval} shows similar trends for~$k \in \{10,100\}$.

\textbf{Framing the Task as Ranking Does Not Help Much.}
Binary judgments are hard in isolation because they~use implicit cutoffs; relative comparisons are often easier \citep{10.1561/1500000016}.
We~thus test alternative formulations of action prediction: (1) jointly labeling all actions per query, mirroring our annotation; and (2) pairwise comparisons over all action pairs---ranking by ELO score with a validation-tuned cutoff.
Neither format significantly beats the user-history baseline (Table~\ref{table:ablations}, \textit{Task Engineering}), so action prediction's difficulty is not just its task format.

\textbf{Takeaway:} Action prediction is tough but tractable, largely improving with user-specific history.
Conversely, extra user features, long-context management, and varied task formats offer limited gains beyond this.
For researchers building user simulators, Table~\ref{table:ablations} shows~behavior from the user mirroring the task is the most predictive signal versus self-reported and extrapolated user context, despite their common use in~simulations \citep{10.1145/3708985}.

\subsection{The Limits of Action Prediction: User Preferences are Far from \textcolor{vampire}{Immortal}} \label{subsection:unstable}

\begin{wrapfigure}{r}{0.45\linewidth}
    \centering
    \vspace{-4ex}
    \includegraphics[width=\linewidth]{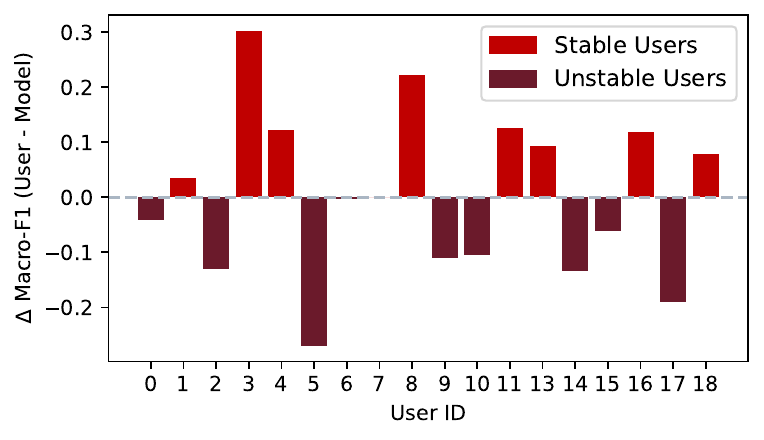}
    \vspace{-4.7ex}
    \caption{\small The Macro-F1 gap between user stability and GPT-5 predictions points to two user types: users with stable preferences GPT cannot reliably capture (e.g., user 3, user 8) and users with less stable preferences where affordances like actions are vital (e.g., user 5).}
    \label{fig:self_consistency}
    \vspace{-1.5ex}
\end{wrapfigure}

Given our improved action prediction scores in \cref{subsection:follow_up_predict}, we now test whether our LLMs are at the limit of what is inherently predictable in \dataset{} by measuring our users' preference stability.
To test this, 18 annotators re-label four of their queries~in the first study, re-selecting the actions they prefer.
We run this five months after the first study to limit memory effects.

User stability---predicting initial user selections from their new ones---has F1=0.672, near GPT-5 on the same subset using user-specific history (0.664), but errors differ in action types~and users.
User stability F1 is closer between~\generic{} and \paper{} actions (0.681 and 0.664, respectively) than GPT (0.638 and 0.689).
Per-user, stability ranges from as low as 0.434 (user 5) to as high as 0.823 (user 3) and 0.945 (user 8) F1.
While GPT-5 improves F1 over less stable users, it struggles in action prediction on stable ones (Figure~\ref{fig:self_consistency})---so there is still headroom for F1 on stable users.

Changes in user selections are not just noise.
Analyzing rationales, one user initially selected the action
``\textit{Interpret hallucinations as generative distribution mismatch problems}'', but later decided not to, as they were now ``\textit{interested in solutions, not new insights}''---indicating a shift in intent.
Similarly, one user initially picked the action ``\textit{Start with concise definitions of fall detection}'' as it was ``\textit{perfect for their audience}'', but later did not since they ``\textit{already know what a fall is}''; their preference changed based on whether they were reading the report (i.e., versus sharing it).

Action prediction is inherently bounded by latent shifts in users' intent and audience \citep{yan-etal-2020-unknown}.
For such users, clarification mechanisms such as our actions remain practical, enabling them to customize report generation in ways models cannot reliably anticipate.

\section{Richer User Contexts Can \textcolor{vampire}{Feed} Personalized Action Generation} \label{subsection:user_context}



We now use our offline results to improve action generation, ``closing the loop''~\citep{humeval-2021-human} by better aligning actions with user preferences.
Our analyses recurringly motivate personalization \citep{Brusilovsky1996MethodsAT}---action generation flaws often stem from user-specific intent (\cref{subsection:coding}) and selection history improves simulation (\cref{subsection:follow_up_predict})---suggesting~action generation can benefit from user-specific context.
However, our first cold-start personalization attempt via papers was subpar (\cref{subsection:data_summary}), so we also test two user contexts that more closely mirror \dr{} interactions: (1) \texttt{stated preferences} from \cref{subsection:follow_up_predict}; and (2) \texttt{revealed preferences} \citep{samuelson1938note}---queries, actions, and judgments users previously provided as few-shot examples.

\begin{wraptable}{r}{0.38\textwidth}
\small
\centering
\vspace{-12pt}
\centering
\begin{tabular}{@{}lc@{}}
\toprule
                     & Selection Rate \\ 
\midrule
Generic/Query Only        & 0.817 ± 0.030 \\ 
Research Papers        & 0.682 ± 0.036 \\
Stated Preferences   & 0.741 ± 0.034 \\
Revealed Preferences & \textbf{0.857 ± 0.027} \\
\bottomrule
\end{tabular}
\vspace{-1.7ex}
\caption{\small Action selection rates across user contexts with 95\% CIs. Generating with user's self-reported ideal \dr{} behavior (stated preferences) and prior action selections (revealed preferences) are stronger than papers for personalization, with the latter selected more often than generic actions (best in \textbf{bold}).}
\label{table:user_context}
\vspace{-1.4ex}
\end{wraptable}

Eighteen annotators write ten more queries to ask our \dr{} system.
For each query, we generate four actions conditioned on each of our four user contexts (\generic{}, \paper{}, \texttt{stated preferences}, \texttt{revealed preferences}) with GPT-4.1---a strong model in Figure~\ref{fig:breakdown}---yielding 16 in total.
Given the variance of selection rates over qualitative action types (\cref{subsection:data_summary}), we no longer enforce type diversity.
GPT-4.1 de-duplicates the actions across contexts.
Annotators then pick actions they would want the \dr{} system to execute with rationales, just like \cref{section:dataset}.

Selection rates for actions from \texttt{stated} and \texttt{revealed preferences} exceed \paper{} actions (Table~\ref{table:user_context}).
Research papers are not tied to \dr{} interactions, highlighting~the challenge of finding useful cold-start personalization signals for \dr{} \citep{zhao2025meta}.
The gains from \texttt{revealed preferences} over \generic{} are modest,~so \generic{} actions already capture broadly useful behavior.
But~crucially, personalized action selection rates do not need to exceed generic actions to be~useful;~personalization yields varied actions---only 12.1\% overlap between \generic{} and \texttt{revealed~preferences}---that users still pick above random (>50\%), expanding the actions \dr{} can take to help~users.

We investigate the benefits of \texttt{revealed preferences} via GPT-4.1's rationales for generated actions, which cite relevant items from the user's past selections.
80.0\% of rationales~cite both generic and paper actions in the user's history, suggesting GPT combines the strengths of varied action types.
One action, ``\textit{Compare CLIP vs other~benchmarked VLMs}'', merged~the~user's preference for comparisons via \generic{} actions (e.g., ``\textit{Present table showing models}'') and the user's interest in benchmarks found in \paper{} actions (e.g., ``\textit{Highlight benchmark evaluation}'').

Overall, the benefits of past selections show the potential for \dr{} to learn from accumulated interaction data \citep{10.1145/1772690.1772758}.
More broadly, it is promising that gains from user-specific context in action prediction correspond to improvements in generated actions.
We encourage future work to explore the extent to which simulation accuracy on real user feedback \citep{zhou2026mind, seshadri2026lost} can inform online interventions to improve agent utility.  

\section{\textcolor{vampire}{Digging} Up Related Work}

Given our paper's focus on collecting intermediate feedback through generated actions in \dr{}, we bridge prior work in \dr{} evaluation (\cref{subsection:deep_research}) and intermediate user feedback (\cref{subsection:feedback}).

\subsection{Evaluating Deep Research Systems} \label{subsection:deep_research}

Scientific Deep Research (\dr{}) agents synthesize long-form reports via documents to aid scientific~research \citep{team2025tongyi}.
By combining retrievers and LLMs \citep{Asai2024OpenScholarSS, wang2024autosurvey, lala2023paperqa}, such agents can create varied report formats~like~Wikipedia articles \citep{liu2018generating}, expository texts \citep{jiang2025archidocgen}, and surveys \citep{yan2025surveyforge}.  

Beyond excelling on benchmarks \citep{wei2025browsecomp, Chandrahasan2025DeepRC}, training and evaluating with user feedback is a gold standard for improving \dr{} \citep{ouyang2022training, touvron2023llama}. 
Most protocols rely on ratings \citep{liwebthinker, hwang2026deep} or preferences \citep{zhao2025sciarena, miroyan2025search, du2026deepresearch} to collect user feedback on reports, but this feedback does not directly target the intermediate actions \dr{} agents take \citep{zhong2026dualspec}---retrieving papers, planning sections, and writing reports---obscuring the actions users prefer.
While some works reconstruct intermediate feedback based on user judgments over the final response \citep{garbacea2025hyperalign, joshi2025improving, findeis2025InverseConstitutionalAI}, these are hypotheses and not confirmed by users \citep{li2025eliciting}.
\dataset{} instead directly collects feedback on actions user prefer for \dr{}, along with their execution quality.

\subsection{Intermediate User Feedback} \label{subsection:feedback}

Researchers have long recognized the weaknesses of user feedback on final outputs \citep{hosking2024human}, and have thus designed protocols to collect more informative signals.
Examples include fine-grained and multi-dimensional feedback \citep{10.5555/3666122.3668696, wang-etal-2024-interpretable}, improved specification in annotation protocols \citep{dipper-etal-2004-towards, 10.1162/TACL.a.24}, and richer signals from downstream interactions \citep{mozannar2025the, shaikh2026learning}.

In \dr{}, intermediate feedback is rarely collected. 
Works design multi-aspect rubrics for~\dr{} queries \citep{lv2026learning, li2026deepresearch, li2026rubrichub, sharma2025researchrubrics, Gunjal2025RubricsAR}, but~they are often designed by experts to score objectively high-quality behaviors, not subjective user preferences.
\citet{balepur2026dont} and \citet{pan2026interdeepresearch} also generate actions for \dr{} but evaluate them only via qualitative interviews, in contrast to our large-scale feedback collection and simulation experiments. 
\citet{Chandrahasan2025DeepRC} release a \dr{} feedback platform and 1,281 votes on tool calls in agent execution traces (e.g., ``Scraped five web pages...''), but this setup mainly evaluates execution quality.  
Conversely, \dataset{} is the first dataset with 8,310 points of feedback for understanding \textit{which} actions users want \dr{} systems to~execute.

Outside of \dr{}, the closest line of work is process reward modeling \citep[PRMs]{lightman2024lets}, which gathers step-level feedback on model reasoning \citep{zhang-etal-2025-lessons} or agent traces \citep{zhang2026webarbiter} to reward strong executions. 
\dataset{}'s action feedback is similar in spirit, but unlike PRMs' focus on improving how a chosen trajectory is executed, our feedback guides which trajectories an agent should follow.
This makes action feedback closer to query reformulation \citep{li2024learning}: actions are evaluated before being executed, while feedback in PRMs typically scales with the number of steps in the execution trace.

\section{Conclusion: Ignoring the Actions Users Want is a \textcolor{vampire}{Grave} Mistake}

We release \dataset{}: a dataset of user feedback on intermediate actions in \dr{}, moving beyond the default protocol of only evaluating \dr{} reports.
Our data shows systems cannot fully anticipate the actions users select for \dr{}, but can execute them well once specified.

Generating useful actions is underexplored, but \dataset{} enables models for predicting ones users want \dr{} to take.
When reliable, such methods can form rewards in new training methods or benchmarks \citep{shao2025dr}.
Toward this, we reveal the dominance of user-specific history for action prediction, motivating new research in user context: managing long action selection history \citep{wang2024crafting}, discovering better signals than papers in cold-start settings \citep{zhao2025meta}, and inferring preferences as text \citep{ryan-etal-2025-synthesizeme}.

Our success in action execution does not imply it is solved.
Challenges remains in improving execution as user requests grow complex \citep{pham-etal-2024-suri}.
The actions users~do~\underline{not}~pick can reveal execution gaps for future work.
While rare in \dataset{}, users avoided some~actions (e.g., ``\textit{Add code snippets}'') they knew the \dr{} system could not execute (\cref{subsection:coding}), showing how action-level user feedback can surface emerging execution needs \citep{haddad2026understanding}.

More broadly, our decomposition offers a scalable way to collect feedback in complex agents: articulating potential actions users can endorse, then executing them in the final output.
Since action generation only requires specifying the types of actions that the end agent can execute, feedback on actions can be collected independently to execution improvements.
As agent horizons grow \citep{task-completion-time-horizons-of-frontier-ai-models}, action-level feedback will become increasingly crucial for making agents more helpful to users.
\dataset{} represents one step toward this goal.



\section*{Ethics Statement}

While learning from human feedback can improve user satisfaction in \dr{}, we acknowledge potential risks---especially when considering personalization and user-specific context.
Optimizing for actions that users select may instill confirmation bias with what they are already familiar with \citep{klayman1995varieties}, raise privacy and stereotyping concerns \citep{kantharuban-etal-2025-stereotype}, and may lead to responses that are users think are helpful without truly improving downstream outcomes \citep{mozannar2025the}.
To help mitigate these risks, we only consider user outputs contexts tied to our system (e.g., research papers, stated preferences) and not sensitive information, and our organization's IRB approved the entire user study design.

We also acknowledge that our results only directly apply to our specific setting of Deep Research with CS researchers.
As the first study on action-level feedback in \dr{} and given~the cost of our user studies, we primarily study CS, following prior work \citep{bragg2025astabench}.
Future work should ensure our results generalize to their desired personalization tasks and user populations before deploying any of the interventions we study in this work.

We use Generative AI in this project.
We use Cursor to help implement experiments and modify the style of plots, and ChatGPT to help refine paper writing.



\section*{Acknowledgments}

We would like to thank the Allen Institute for Artificial Intelligence (Ai2), the CLIP lab at the University of Maryland, and external collaborators for their help.
We appreciate discussions with Hung-Ting Chen, Varun Yerram, Ayush Jhaveri, Deniz Qian, Dan Weld, Matt Latzke, Jonathan Bragg, Jay DeYoung, and Jena D. Hwang on earlier versions of our paper.
In particular, we thank Dang Nguyen and Yapei Chang for both supporting and critiquing the \textcolor{vampire}{vampire} puns in this paper.

This material is based upon work supported by the National Science Foundation under \abr{iis}-2339746 (Rudinger) \abr{iis}-2403436 (Boyd-Graber), and \abr{dge}-2236417 (Balepur).
Any opinions, findings, and conclusions or recommendations expressed in this material are those of the author(s) and do not necessarily reflect the views of the National Science Foundation.

\bibliography{colm2026_conference}
\bibliographystyle{colm2026_conference}

\clearpage

\appendix

\section{Appendix}

\subsection{Model Configurations} \label{appendix:section:model}

For action generation in data collection and user studies (\cref{section:dataset}, \cref{section:generation}), we use the following model endpoints in LiteLLM:
\begin{itemize}
    \item GPT-4.1 \citep{achiam2023gpt}: openai/gpt-4.1-2025-04-14
    \item Gemini-2.5 Flash \citep{comanici2025gemini}: gemini/gemini-2.5-flash
    \item Claude-4 Sonnet \citep{anthropic_claude4_2025}: anthropic/claude-sonnet-4-20250514
    \item Deepseek-V3 \citep{liu2024deepseek}: together\_ai/deepseek-ai/DeepSeek-V3
\end{itemize}

We generate with default parameters. To incentivize diversity during data collection, we generate twice as many actions (e.g., eight generic actions, eight paper-based actions), then randomly sample four of each action type, similar to overgenerate-and-rerank pipelines \citep{hu-etal-2024-reranking}.
We detect duplicate actions between the generic and paper-based action types with GPT-4.1 (Prompt~\ref{prompt:deduplication}), giving credit to both action types if selected, and resampling another random action to ensure the total number of actions reaches eight.
When experimenting with different user contexts in \cref{subsection:user_context}, we only generate four actions of each type and run the same de-duplication process.

For action prediction and action execution experiments (\cref{section:experiments}), we use the following model endpoints in LiteLLM:
\begin{itemize}
    \item Claude-4.6 Opus \citep{anthropic_claude_opus_4_6_2026}: anthropic/claude-opus-4-6
    \item Gemini-3 Pro \citep{google_gemini3_2025}: gemini/gemini-3-pro-preview
    \item GPT-5 \citep{singh2025openai}: openai/gpt-5-2025-08-07
    \item DeepSeek-R1 \citep{guo2025deepseek}: together\_ai/deepseek-ai/DeepSeek-R1
    \item Qwen-3 235B Thinking \citep{yang2025qwen3}: together\_ai/Qwen/Qwen3-235B-A22B-Thinking-2507
\end{itemize}

We generate with a temperature of $1.0$, reasoning effort set to ``medium'' (or equivalents), and a maximum token length of 81920 (but override this for models with lower maximum token lengths).

For Deep Research report generation, we adapt the open-source \modelFull{} pipeline from Ai2 \citep{singh-etal-2025-ai2}.
The system exposes functions for rewriting queries, planning an outline of section titles, and generating sections iteratively.
We minimally modify prompts in these functions so that they also have access to all of the user's selected actions, instead of just the query.
Our preliminary testing found that this led to slightly higher instruction-following versus concatenating all actions with the input query.
However, the later still was effective, so it is feasible for future work to run action feedback collection similar to ours but treating the end agent as a fully black box without modifying any parts of the agent trace.

\subsection{User Interface} \label{appendix:section:user_interface}

\begin{figure}
    \centering
    \includegraphics[width=\linewidth]{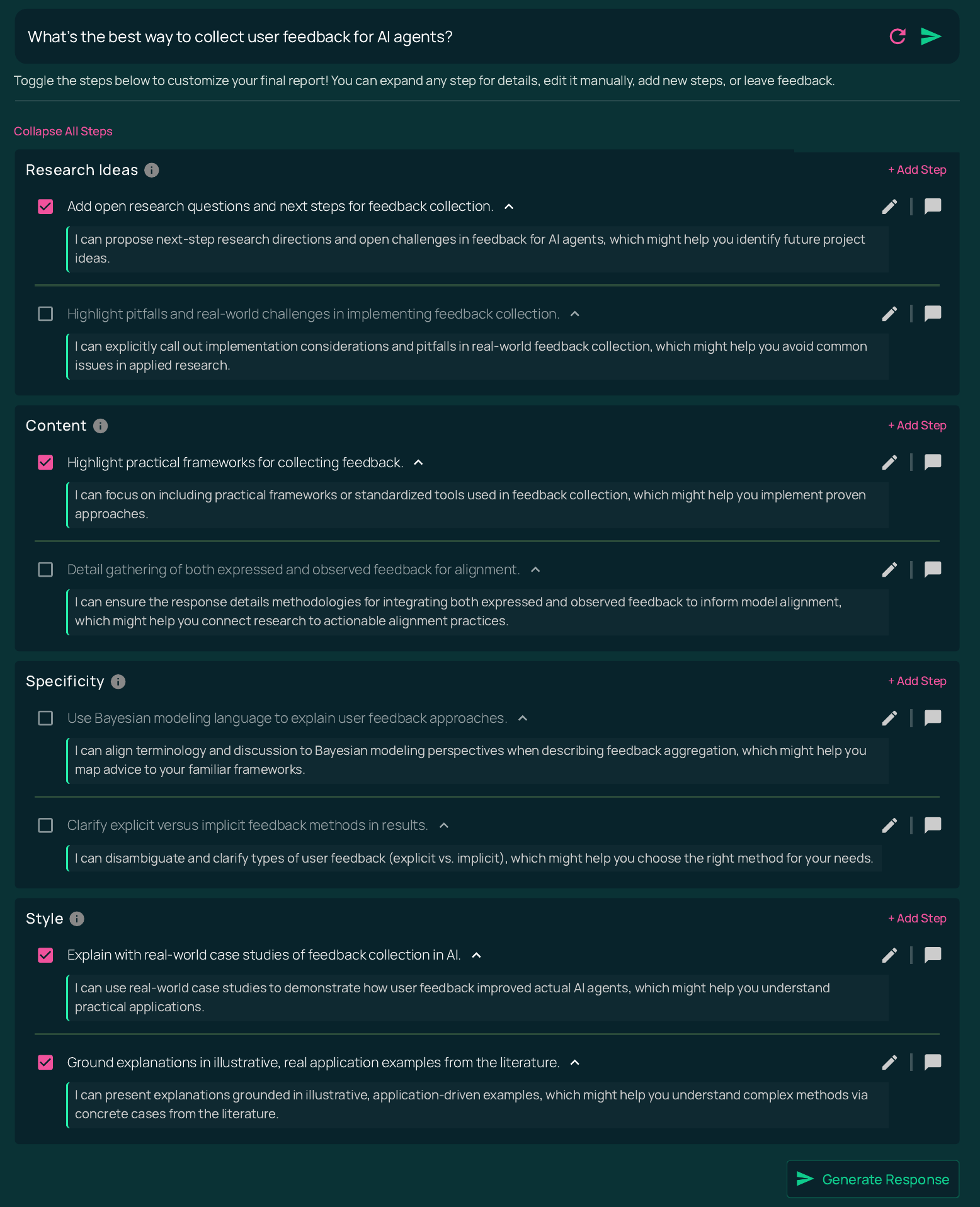}
    \caption{\small Interface overview for action generation in \dataset{}. After annotators issue a query, they see a set of actions---high-level decisions on how the \dr{} system could construct the report. Actions are grouped by how they will impact the report qualitatively. Our annotators select the actions they personally want the system to take and provide a rationale, forming action selection feedback.}
    \label{appendix:fig:plan}
\end{figure}

\begin{figure}
    \centering
    \includegraphics[width=\linewidth]{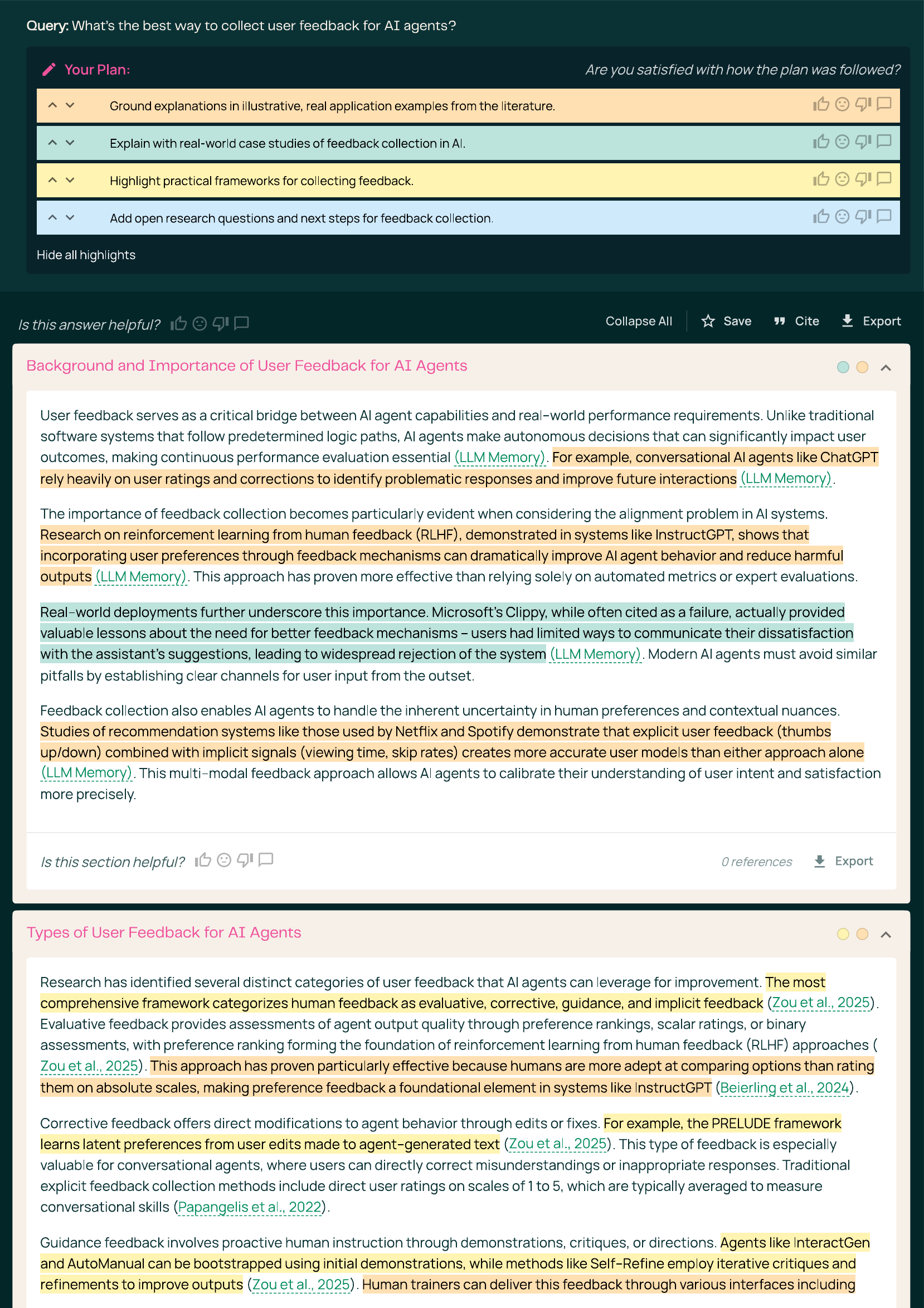}
    \caption{\small Interface overview for action execution in \dataset{}. Given the query and actions selected by the annotator, we query \modelFull{} to generate a report. Annotators read the report and judge whether the system executed each action well (upvote) or poorly (downvote or neutral vote) based on their own personal satisfaction and a rationale, forming action execution feedback. To facilitate this annotation, we modify \modelFull{} to generate highlights showing where each action was executed.}
    \label{appendix:fig:report}
\end{figure}

We provide screenshots of the interface annotators use for evaluating action generation (Figure~\ref{appendix:fig:plan} and action execution (Figure~\ref{appendix:fig:report}).
The interface was built by extending the companion web app to the \modelFull{} system \citep{singh-etal-2025-ai2}.
The interface was built by extending the companion web app to the \modelFull{} system \citep{singh-etal-2025-ai2}.

\subsection{Further Comparisons of Action and Execution Prediction} \label{appendix:section:compare_prediction}

Along with action prediction, we also experiment with execution prediction---where LLMs must predict whether a user judged an action as being executed well (0/1) given the report, query, and action.
We provide further results on action prediction and execution prediction across more LLMs, metrics, zero-shot and few-shot prompting strategies, and action types in Table~\ref{tab:action_prediction} and Table~\ref{tab:report_prediction_all_users}, respectively.
Our results support \cref{subsection:initial_predict} across these settings: action prediction largely improves when models access user-specific history, but the task still has lower scores than execution prediction.

\subsection{Further Evaluations of Action Prediction with Retrieval} \label{appendix:section:retrieval}

We further explore the benefits of retrieval for long-context management in action prediction.
We replicate our experiments in \cref{subsection:follow_up_predict} with $k \in \{10, 100\}$ and introduce two new retrieval baselines: retrieving the $k$-oldest examples from the user's history and the $k$-newest examples.
Overall, our findings are consistent (Table~\ref{appendix:table:retrieval}): retrieval minimally improves F1 scores in some cases, but the fact that retrieval even with $k=10$ does not largely degrade scores shows there is room to improve efficiency for user modeling in the task.

\subsection{Further Evaluations of Action Prediction with Stated Preferences} \label{appendix:section:stated_preferences}

We explore how well other LLMs can infer rules from a user's action history and apply them for predictions, following the procedure of \textit{Stated Preferences} in \cref{subsection:follow_up_predict}.
Overall, trends are consistent: user-specific history dominates in F1, and model-inferred rules are more useful for predictions than user-stated rules.

\subsection{Clustering Procedure} \label{appendix:section:clustering}

We now detail the qualitative clustering procedure used in \cref{subsection:coding}.
Our process is inspired by qualitative coding \citep{bingham2023data}.
We first prompt Claude-4.6 Sonnet to use 200 random examples from each user's history to derive high-level clusters describing either actions (Prompt~\ref{prompt:cluster_generate_action}) or rationales (Prompt~\ref{prompt:cluster_generate_rationale}).
We then prompt the model to merge all per-user clusters into higher-level themes, forming a global taxonomy (Prompt~\ref{prompt:cluster_merge}).
The first and last author reviewed all clusters and discarded low-quality ones (e.g., ``Add Specific Content Section'' for action clusters, as it overlaps with categories like ``Restructure Organization''). 

Finally, we have Gemini-3-Flash take each individual action and rationale and classify which cluster it belongs to (Prompt~\ref{prompt:cluster_label} with temperature equal to $0.0$.
One author reviewed 50 random classification outputs across 500 examples for action and rationale labeling. 
They agreed with the classifier in 98\% of cases for action clustering and 90\% of cases for rationale clustering.

\subsection{Prompts} \label{appendix:section:prompts}

To encourage reproducibility during the review process, we provide the prompts used in our experiments here, but encourage readers to access our Github upon release to access them more easily.

For action generation, we provide prompts for generic action generation (Prompt~\ref{prompt:action_generation_generic}), paper-based action generation (Prompt~\ref{prompt:action_generation_paper}), rule-based action generation (Prompt~\ref{prompt:action_generation_stated}), history-based action generation (Prompt~\ref{prompt:action_generation_revealed}).
During data collection, the generic and paper-based prompts must generate a fixed number of actions and evenly distributed to the following qualitative action types:

\begin{itemize}
  \item \textbf{Content}: Specifies {what information} the response should include and how it should be conceptually framed.
  \item \textbf{Explanation Style}: Specifies the {style of the response} and {how the explanation for the information is communicated}.
  \item \textbf{Specificity}: Clarifies and narrows the scope of the response to better match the researcher's intended focus.
  \item \textbf{Usefulness (Research Ideas)}: Shapes the response to be {actionable} or {instrumental} for the researcher's goals or workflow.
\end{itemize}

and the following implementation types, which are given to different steps in \model{} so it knows how to alter reports:

\begin{itemize}
  \item \textbf{Search Add}: Personalizes the search by {adding new terms or dimensions} to the original query.
  \item \textbf{Search Refine}: Personalizes the search by {revising or improving} the original query.
  \item \textbf{Organization}: Personalizes how the papers are grouped into sections for the final response.
  \item \textbf{Generation}: Personalizes how certain sections are written and explained.
\end{itemize}

During the user study in \cref{subsection:user_context}, we do not fix the distribution of the qualitative action types.
We also provide our prompts for de-duplication (Prompt~\ref{prompt:deduplication}).

For clustering, we provide our generation prompts for action clusters (Prompt~\ref{prompt:cluster_generate_action}) and rationale clusters (Prompt~\ref{prompt:cluster_generate_rationale}), merging clusters per user (Prompt~\ref{prompt:cluster_merge}), and labeling new actions and rationales according to existing cluster labels (Prompt~\ref{prompt:cluster_label}).

Finally, we provide action prediction prompts in binary classification (Prompt~\ref{prompt:action_prediction}), ranking (Prompt~\ref{prompt:action_multi_classification}), and pairwise comparison (Prompt~\ref{prompt:action_ranking}) task formats. We also provide our action execution prompt (Prompt~\ref{prompt:action_execution}) and prompt for inferring rules from the user's action selection history (Prompt~\ref{prompt:rule_inference}).
We tailor these prompts for all of the experiments in \cref{subsection:follow_up_predict}, such as when we add new features.

\clearpage

\begin{table}[]
\small
\scriptsize
\setlength{\tabcolsep}{3.5pt}
\begin{tabular}{@{}lcccc|cccc|cccc@{}}
\toprule
  & \multicolumn{4}{c}{\textit{Generic}} & \multicolumn{4}{c}{\textit{Papers}} & \multicolumn{4}{c}{\textit{Overall}} \\
\cmidrule(lr){2-5} \cmidrule(lr){6-9} \cmidrule(lr){10-13}
Model & Acc & F1 & $\kappa$ & MCC & Acc & F1 & $\kappa$ & MCC & Acc & F1 & $\kappa$ & MCC \\ \midrule
C-4.6 Opus (0-shot) & 0.6321 & 0.5784 & 0.1650 & 0.1688 & 0.5935 & 0.5918 & 0.2459 & 0.2904 & 0.6176 & 0.6036 & 0.2322 & 0.2486 \\
C-4.6 Opus (All History) & 0.6749 & 0.5698 & 0.1419 & 0.1430 & 0.6908 & 0.6819 & 0.3657 & 0.3681 & 0.6895 & 0.6405 & 0.2811 & 0.2811 \\
C-4.6 Opus (User History) & 0.7443 & 0.6810 & 0.3620 & 0.3620 & 0.7318 & 0.7260 & 0.4554 & 0.4612 & 0.7388 & 0.7069 & 0.4147 & 0.4162 \\
\midrule
G-3 Pro (0-shot) & 0.5756 & 0.5514 & 0.1452 & 0.1614 & 0.5461 & 0.5339 & 0.1904 & 0.2614 & 0.5610 & 0.5587 & 0.1874 & 0.2231 \\
G-3 Pro (All History) & 0.7043 & 0.5376 & 0.1071 & 0.1203 & 0.6559 & 0.6133 & 0.2370 & 0.2443 & 0.6917 & 0.5901 & 0.1988 & 0.2116 \\
G-3 Pro (User History) & 0.7308 & 0.6247 & 0.2559 & 0.2622 & 0.7268 & 0.7025 & 0.4076 & 0.4114 & 0.7344 & 0.6725 & 0.3485 & 0.3532 \\
\midrule
GPT-5 (0-shot) & 0.6659 & 0.6041 & 0.2110 & 0.2128 & 0.6160 & 0.6160 & 0.2688 & 0.2973 & 0.6405 & 0.6211 & 0.2574 & 0.2687 \\
GPT-5 (All History) & 0.6524 & 0.5623 & 0.1245 & 0.1245 & 0.6683 & 0.6665 & 0.3471 & 0.3627 & 0.6678 & 0.6345 & 0.2723 & 0.2750 \\
GPT-5 (User History) & 0.7195 & 0.6416 & 0.2834 & 0.2836 & 0.7343 & 0.7271 & 0.4562 & 0.4597 & 0.7246 & 0.6876 & 0.3755 & 0.3761 \\
\midrule
DS-R1 (0-shot) & 0.6546 & 0.5916 & 0.1863 & 0.1880 & 0.5910 & 0.5910 & 0.2213 & 0.2447 & 0.6264 & 0.6072 & 0.2313 & 0.2423 \\
DS-R1 (All History) & 0.6817 & 0.5145 & 0.0562 & 0.0616 & 0.6234 & 0.5807 & 0.1702 & 0.1742 & 0.6580 & 0.5583 & 0.1300 & 0.1353 \\
DS-R1 (User History) & 0.7059 & 0.5308 & 0.0977 & 0.1118 & 0.6692 & 0.6009 & 0.2353 & 0.2621 & 0.6918 & 0.5728 & 0.1746 & 0.1932 \\
\midrule
Q-3 235B (0-shot) & 0.6953 & 0.5594 & 0.1329 & 0.1394 & 0.6459 & 0.6375 & 0.2786 & 0.2815 & 0.6786 & 0.6229 & 0.2461 & 0.2463 \\
Q-3 235B (All History) & 0.6975 & 0.5166 & 0.0724 & 0.0835 & 0.6459 & 0.5926 & 0.2031 & 0.2140 & 0.6819 & 0.5663 & 0.1585 & 0.1728 \\
Q-3 235B (User History) & 0.7081 & 0.5519 & 0.1293 & 0.1418 & 0.7018 & 0.6675 & 0.3421 & 0.3503 & 0.7115 & 0.6231 & 0.2593 & 0.2715 \\
\bottomrule
\end{tabular}
\caption{\small Action prediction results across models, action types, and three prompt types: zero-shot, few-shot with all user history, and few-shot with user-specific history. Across models and metrics, the trend in \cref{subsection:initial_predict} remains consistent: model predictions largely improve after adding user-specific history.}
\label{tab:action_prediction}
\end{table}



\begin{table}[]
\small
\scriptsize
\setlength{\tabcolsep}{4.3pt}
\begin{tabular}{@{}lcccc|cccc|cccc@{}}
\toprule
  & \multicolumn{4}{c}{\textit{Generic}} & \multicolumn{4}{c}{\textit{Papers}} & \multicolumn{4}{c}{\textit{Overall}} \\
\cmidrule(lr){2-5} \cmidrule(lr){6-9} \cmidrule(lr){10-13}
Model & Acc & F1 & $\kappa$ & MCC & Acc & F1 & $\kappa$ & MCC & Acc & F1 & $\kappa$ & MCC \\ \midrule
C-Opus 4.6 (0-shot) & 0.7305 & 0.6995 & 0.4142 & 0.4390 & 0.7054 & 0.6996 & 0.4150 & 0.4385 & 0.7172 & 0.6958 & 0.4084 & 0.4344 \\
C-Opus 4.6 ($n$-shot) & 0.7143 & 0.6913 & 0.3898 & 0.3997 & 0.6977 & 0.6962 & 0.3976 & 0.4046 & 0.7109 & 0.6973 & 0.4021 & 0.4127 \\
\midrule
G-3 Pro (0-shot) & 0.7208 & 0.6816 & 0.3864 & 0.4215 & 0.6938 & 0.6813 & 0.3935 & 0.4376 & 0.7093 & 0.6795 & 0.3867 & 0.4276 \\
G-3 Pro ($n$-shot) & 0.7435 & 0.7190 & 0.4477 & 0.4653 & 0.6977 & 0.6917 & 0.3996 & 0.4222 & 0.7220 & 0.7037 & 0.4205 & 0.4412 \\
\midrule
GPT-5 (0-shot) & 0.7403 & 0.7215 & 0.4478 & 0.4564 & 0.7093 & 0.7065 & 0.4215 & 0.4341 & 0.7283 & 0.7149 & 0.4375 & 0.4501 \\
GPT-5 ($n$-shot) & 0.7273 & 0.7086 & 0.4216 & 0.4285 & 0.7209 & 0.7166 & 0.4454 & 0.4658 & 0.7267 & 0.7131 & 0.4340 & 0.4470 \\
\midrule
DS-R1 (0-shot) & 0.6818 & 0.6128 & 0.2817 & 0.3434 & 0.6498 & 0.6243 & 0.3069 & 0.3688 & 0.6677 & 0.6139 & 0.2877 & 0.3521 \\
DS-R1 ($n$-shot) & 0.6948 & 0.6402 & 0.3195 & 0.3670 & 0.6202 & 0.5900 & 0.2502 & 0.3075 & 0.6651 & 0.6147 & 0.2846 & 0.3417 \\
\midrule
Q-3 235B (0-shot) & 0.7338 & 0.7051 & 0.4233 & 0.4451 & 0.6744 & 0.6627 & 0.3548 & 0.3899 & 0.7093 & 0.6847 & 0.3900 & 0.4201 \\
Q-3 235B ($n$-shot) & 0.7240 & 0.6908 & 0.3986 & 0.4250 & 0.6744 & 0.6606 & 0.3552 & 0.3967 & 0.6998 & 0.6698 & 0.3671 & 0.4045 \\
\bottomrule
\end{tabular}
\caption{\small Execution prediction results across models, action types, and two prompt types: zero-shot and few-shot with six examples. For all models, the trend in \cref{subsection:initial_predict} remains consistent: models are stronger in execution prediction versus action prediction.}
\label{tab:report_prediction_all_users}
\end{table}
\setlength{\fboxsep}{1pt}

\begin{table}[t]
\small
\centering
\begin{tabular}{@{}rlccc@{}}
\toprule
\multicolumn{1}{l}{} Experiment Design Choice &
Configuration &
\multicolumn{1}{l}{F1 (Generic)} &
\multicolumn{1}{l}{F1 (Paper)} &
\multicolumn{1}{l}{F1 ($\text{Avg}_\text{macro}$)} \\
\midrule

\textit{Baseline}
& \cellcolor{red!20} User-Specific History
& \cellcolor{red!20} 0.6400
& \cellcolor{red!20} 0.7257
& \cellcolor{red!20} 0.6829 \\
\midrule

\multirow{5}{*}{\textit{\begin{tabular}[c]{@{}r@{}}Long-Context\\ ($k=10$)\end{tabular}}}
& Oldest Examples
& 0.6008 & 0.6544 & 0.6276 \\
& Newest Examples
& 0.5882 & 0.6809 & 0.6346 \\
& Random Sample
& 0.6168 & 0.6450 & 0.6309 \\
& GritLM Retriever
& 0.6221 & 0.6606 & 0.6414 \\
& GPT-5 Retriever
& 0.6257 & \textbf{0.7442} & \textbf{0.6850} \\ \cmidrule(l){2-5}

\multirow{5}{*}{\textit{\begin{tabular}[c]{@{}r@{}}Long-Context\\ ($k=50$)\end{tabular}}}
& Oldest Examples
& 0.6061 & 0.6527 & 0.6294 \\
& Newest Examples
& 0.6185 & 0.6960 & 0.6573 \\
& Random Sample
& 0.6202 & 0.6622 & 0.6412 \\
& GritLM Retriever
& 0.6400 & 0.7105 & 0.6753 \\
& GPT-5 Retriever
& 0.6221 & 0.7061 & 0.6641 \\
\cmidrule(l){2-5}

\multirow{5}{*}{\textit{\begin{tabular}[c]{@{}r@{}}Long-Context\\ ($k=100$)\end{tabular}}}
& Oldest Examples
& \textbf{0.6460} & 0.6761 & 0.6611 \\
& Newest Examples
& 0.6201 & 0.7087 & 0.6644 \\
& Random Sample
& 0.6326 & 0.6839 & 0.6583 \\
& GritLM Retriever
& 0.6166 & 0.7202 & 0.6684 \\
& GPT-5 Retriever
& 0.6185 & 0.7053 & 0.6619 \\

\bottomrule
\end{tabular}
\caption{\small Extended action prediction retrieval results for $k \in \{10, 50, 100\}$ when using the oldest examples, newest examples, random sampling, and retrieval from either GritLM or GPT-5.
Our results mirror \cref{subsection:follow_up_predict}: retrieval brings minimal gains, but shows user history can be made more efficient.
}
\label{appendix:table:retrieval}
\end{table}
\setlength{\fboxsep}{1pt}

\begin{table}[t]
\small
\centering
\begin{tabular}{@{}rlccc@{}}
\toprule
\multicolumn{1}{l}{} Experiment Type &
Configuration &
\multicolumn{1}{l}{F1 (Generic)} &
\multicolumn{1}{l}{F1 (Paper)} &
\multicolumn{1}{l}{F1 ($\text{Avg}_\text{macro}$)} \\
\midrule

\textit{Baseline}
& \cellcolor{red!20} User-Specific History
& \cellcolor{red!20} \textbf{0.6400}
& \cellcolor{red!20} \textbf{0.7257}
& \cellcolor{red!20} \textbf{0.6829} \\
\midrule

\multirow{2}{*}{\textit{GPT-5}}
& User-Stated Rules
& 0.5860
& 0.6284
& 0.6072 \\
& Model-Inferred Rules
& 0.5852
& 0.6760
& 0.6306 \\
\cmidrule(l){2-5}

\multirow{2}{*}{\textit{Claude-4.6 Opus}}
& User-Stated Rules
& 0.5850
& 0.6048
& 0.5949 \\
& Model-Inferred Rules
& 0.5983
& 0.6570
& 0.6277 \\
\cmidrule(l){2-5}

\multirow{2}{*}{\textit{Gemini-3 Pro}}
& User-Stated Rules
& 0.5739
& 0.5804
& 0.5772 \\
& Model-Inferred Rules
& 0.6064
& 0.6294
& 0.6179 \\

\bottomrule
\end{tabular}

\caption{\small Comparison of user-specific history with user-stated and model-inferred preference rules when using different generator and predictor models for action prediction. No combination can surpass using the full user-specific history and model-inferred rules are more useful than user-stated rules for action predictions.}
\label{appendix:table:stated_preferences}
\end{table}

\clearpage
\scriptsize
\centering
\captionsetup{singlelinecheck=false}
\setlength{\LTleft}{\fill}
\setlength{\LTright}{\fill}
\renewcommand{\arraystretch}{1.2}
\setlength{\tabcolsep}{4pt}
\rowcolors{2}{gray!8}{white}
\begin{longtable}{>{\raggedright\arraybackslash}p{0.20\textwidth} >{\raggedright\arraybackslash}p{0.50\textwidth} >{\centering\arraybackslash}p{0.11\textwidth} >{\centering\arraybackslash}p{0.11\textwidth}}
\caption{The clusters uncovered when qualitatively organizing actions with Claude-4.6 Opus, along with selection rates and variance of selection rates across users. The descriptions are the exact ones generated by Claude. \label{appendix:table:action_cluster_full}}\\
\toprule
Cluster title & Cluster description & $\mu$ Action Sel. & User $\sigma^2$ \\
\midrule
\endfirsthead
\toprule
Cluster title & Cluster description & $\mu$ Action Sel. & User $\sigma^2$ \\
\midrule
\endhead
\bottomrule
\endfoot
Summarize Key Findings (N=332) & Adding concise summaries, bullet-point takeaways, executive overviews, key finding syntheses, evaluative digests, or scannable enumerated lists to help readers rapidly grasp the report's core messages without reading full detail. & 0.798 & 0.041 \\
Comparison Analysis (N=461) & Adding structured comparison tables, matrices, side-by-side evaluations, or narrative comparative analyses between methods, approaches, technologies, or systems to help readers assess relative strengths, differences, and trade-offs. & 0.774 & 0.026 \\
Restructure Organization (N=364) & Reorganizing the report's overall structure, section ordering, layout, or presentation framework — including adopting specific document types (e.g., blog post, FAQ, technical report, survey), using structural patterns (e.g., problem-solution, chronological, taxonomy-based), adding navigation aids, or reformatting sections for improved flow and readability. & 0.772 & 0.020 \\
Add Ethical Regulatory Content (N=101) & Including dedicated content on ethical considerations, regulatory compliance frameworks, standardization efforts, data privacy, societal impact, or responsible deployment concerns related to the technologies or methods discussed. & 0.752 & 0.117 \\
Add Concrete Examples (N=315) & Incorporating concrete real-world examples, case studies, worked walkthroughs, illustrative scenarios, before-and-after demonstrations, or qualitative illustrations to make abstract concepts tangible, accessible, and grounded in practice. & 0.746 & 0.025 \\
Add Historical Context (N=79) & Introducing historical context, origin stories, evolutionary timelines, or paradigm-shift narratives showing how techniques, architectures, or fields have developed over time. & 0.734 & 0.042 \\
Include Empirical Evidence (N=299) & Prioritizing or adding quantitative benchmarks, experimental data, performance metrics, empirical results, direct quotes from studies, or evidence-backed findings to ground the report in concrete, measurable evidence rather than theoretical or speculative content. & 0.722 & 0.030 \\
Tone And Narrative Style (N=261) & Modifying the report's writing style, tone, voice, language register, or narrative structure — including adopting assertive, formal, conversational, or engaging tones; using storytelling or problem-solution narratives; adjusting audience expertise level; or calibrating vocabulary for accessibility or domain specificity. & 0.686 & 0.037 \\
Define And Explain Concepts (N=455) & Adding definitions of key terms, clarifying terminology, providing analogies or intuitive explanations, coining memorable labels, simplifying language for accessibility, disambiguating ambiguous concepts, or offering conceptual explanations to ensure reader comprehension of technical or abstract material. & 0.686 & 0.029 \\
Critical Analysis And Limitations (N=263) & Emphasizing critical evaluation of methods including limitations, trade-offs, failure modes, common pitfalls, cost-benefit considerations, challenges, and balanced assessments of strengths versus weaknesses to provide a realistic and nuanced perspective. & 0.677 & 0.055 \\
Practical Guidance (N=544) & Adding actionable implementation advice, step-by-step guides, checklists, decision frameworks, troubleshooting tips, deployment recommendations, code snippets, policy recommendations, or practical workflow details to help readers directly apply or operationalize findings. & 0.675 & 0.053 \\
Add Visualizations And Diagrams (N=49) & Including figures, diagrams, flowcharts, architectural illustrations, workflow visuals, or other graphical elements to clarify complex structures, processes, or relationships. & 0.673 & 0.116 \\
Narrow Scope And Exclude Content (N=772) & Restricting, narrowing, or filtering the report's content coverage by excluding irrelevant topics, limiting to specific subtopics or domains, skipping basic or introductory material, or assuming reader expertise — any action that reduces breadth to increase focus and relevance. & 0.662 & 0.020 \\
Add Technical Depth (N=466) & Increasing the report's technical rigor by adding mathematical formulations, formal derivations, algorithmic details, architectural specifics, computational complexity analyses, hyperparameter descriptions, methodological walkthroughs, or theoretical foundations. & 0.639 & 0.047 \\
Evaluation Framework (N=264) & Providing systematic evaluation frameworks, summarizing available datasets and benchmarks, describing evaluation metrics and protocols, or adding novel measurement approaches to support rigorous assessment of methods. & 0.629 & 0.055 \\
Add Faq Section (N=81) & Appending a dedicated FAQ or question-and-answer section to the report to address anticipated reader questions, common concerns, or misconceptions in a structured Q\&A format. & 0.617 & 0.124 \\
Refine Literature Search (N=629) & Adjusting the literature search strategy by modifying search terms, filtering by paper type or quality criteria, prioritizing specific sources (e.g., recent, high-impact, empirical, foundational), broadening to adjacent areas, or increasing citation density to improve the relevance and quality of the report's evidence base. & 0.604 & 0.049 \\
Future Directions And Gaps (N=489) & Including open research questions, identified gaps in the literature, limitations of current work, future research directions, follow-up suggestions, proposed experiments, or next steps that orient the reader toward unsolved problems and emerging opportunities. & 0.599 & 0.041 \\
Shift Thematic Focus (N=765) & Reinterpreting or redirecting the report's thematic emphasis — reframing the query intent, prioritizing specific capabilities or perspectives (e.g., security, equity, scalability), highlighting a particular method category or application domain, tailoring to a specific real-world context, or foregrounding a particular analytical angle without fully excluding other content. & 0.597 & 0.035 \\
Add Tools And Resources (N=128) & Including references to specific tools, code repositories, open-source software, datasets, simulation frameworks, pre-trained models, or other reusable artifacts to support reproducibility, practical work, and hands-on exploration. & 0.586 & 0.097 \\
Cross Domain Connections (N=105) & Drawing explicit connections between the primary topic and related fields, incorporating cross-disciplinary insights, or linking concepts across domains to broaden context and suggest novel cross-pollination opportunities. & 0.486 & 0.095 \\
\end{longtable}
\rowcolors{2}{}{}
\normalsize

\clearpage
\scriptsize
\centering
\captionsetup{singlelinecheck=false}
\setlength{\LTleft}{\fill}
\setlength{\LTright}{\fill}
\renewcommand{\arraystretch}{1.2}
\setlength{\tabcolsep}{6pt}
\rowcolors{2}{gray!8}{white}
\begin{longtable}{>{\raggedright\arraybackslash}p{0.27\textwidth} >{\raggedright\arraybackslash}p{0.67\textwidth}}
\caption{Uncovered reasons that actions were not selected by participants based on qualitative analyses of their rationales with Claude-4.6 Opus, sorted by N descending. The descriptions are the exact ones generated by Claude. \label{appendix:table:bad_rationale_cluster_full}}\\
\toprule
Cluster title & Cluster description \\
\midrule
\endfirsthead
\toprule
Cluster title & Cluster description \\
\midrule
\endhead
\bottomrule
\endfoot
Scope Too Narrow Or Excludes Content (N=394) & The action restricts, narrows, filters, prioritizes, or excludes content more than the user wants. The user seeks broader, more general, or more inclusive coverage and objects to the action confining the work to a specific subset, method, domain, time period, or modality. This also includes actions that propose skipping or omitting foundational content, removing material the user wants to retain, or imposing constraints the user considers unnecessary — any case where the action would cause the user to lose breadth or relevant material. \\
Topic Not Relevant (N=319) & The action addresses a topic, domain, or subject matter that is entirely unrelated to the user's query. There is no meaningful connection between the action and the user's goal — it is simply off-topic, and the user sees no reason it was suggested. \\
Misaligned With User Intent (N=293) & The action is in the right general area but targets the wrong aspect, angle, interpretation, or goal. The system misread the user's specific focus, misidentified what dimension of the topic matters, or pursued a direction that conflicts with the user's stated purpose. The action is not off-topic but is pointed in the wrong direction relative to what the user actually wants. \\
Exceeds Needed Scope (N=257) & The action expands the scope beyond what the user needs, introducing unnecessary breadth, tangential content, follow-up directions, future research questions, or supplementary sections that would dilute focus, bloat the output, or distract from the core objective. The user has a well-defined need and the action overshoots it, adding content the user did not request and does not want. \\
Unnecessary Or Low Value (N=224) & The user considers the action not needed, not worth the effort, or of insufficient value for their current objective. The action may be tangentially on-topic but is superfluous, optional, adds marginal benefit, or is something the user can handle without explicit system help. This includes generic 'not needed' dismissals, acknowledged-but-deprioritized content, skepticism about the action's practical payoff, and cases where the user trusts the system to handle it implicitly without a directive. \\
Wrong Level Of Depth Or Abstraction (N=204) & The action operates at the wrong level of technical depth, granularity, or abstraction relative to the user's needs. Most commonly, the action is too implementation-focused, too mathematically detailed, too procedural, or too advanced when the user wants conceptual, high-level, or survey-level understanding — or vice versa. This includes miscalibration of expertise assumptions (assuming too much or too little background) and actions that would overcomplicate the output beyond what the question warrants. \\
Wrong Format Tone Or Style (N=193) & The user objects to the proposed presentation format, structural organization, writing tone, rhetorical devices, or communication style — not the underlying substance. This includes rejections of inappropriate formats (FAQ, tutorial, blog, technical report), unwanted analogies or simplified framing that risks oversimplification, wrong tone (too formal, too casual, too assertive), non-standard terminology when technical precision is preferred, forced examples or illustrative devices, and structural approaches that clash with the user's expectations or the topic's complexity. \\
Positive Endorsement (N=95) & The user's rationale is actually an affirmative endorsement explaining why the action IS valuable, relevant, or well-aligned with their goals. These are not genuine rejection rationales — they represent selected or approved actions where the user provides supportive reasoning such as 'this is relevant', 'this is valuable', or 'I added this because...' \\
Factually Incorrect Or Inapplicable (N=75) & The action is based on a factual error, a misunderstanding of the domain or concept, an inapplicable premise, incorrect technical framing, or a nonsensical assumption. The user believes the action is wrong, misleading, hallucinated, or built on a flawed understanding of the subject matter — including cases where the system fundamentally misunderstands a key term or concept in the query. \\
Premature Or Out Of Sequence (N=72) & The action may have relevance but is not appropriate for the user's current stage of work or inquiry. The user defers it using temporal language ('right now', 'at this stage', 'later') signaling that the action is premature, out of sequence, or belongs to a future phase of their workflow — not that it is fundamentally wrong. \\
Already Known Or Unnecessary For Audience (N=66) & The user (or their target audience) already possesses the knowledge the action would provide. The action offers definitions, introductions, or foundational content that is redundant given existing expertise, making it a waste of space or patronizingly basic. \\
Unclear Or Underspecified Action (N=51) & The action is too vague, poorly described, confusing, overly generic, or underspecified for the user to evaluate or act on. The user cannot determine what the action would produce, how it connects to their query, or why it would be helpful. This includes actions that are too open-ended to be actionable, use unclear terminology, or have ambiguous or contradictory framing that prevents meaningful assessment. \\
Profile Driven Irrelevance (N=49) & The user identifies that the action was generated based on their research profile, past interactions, saved interests, or background expertise rather than the actual current query. The action may be topically related to the user's general research area but is contextually irrelevant to what they are asking right now. The user explicitly flags the profile-based or context-biased origin as the reason for rejection. \\
Redundant With Existing Content (N=45) & The action duplicates or substantially overlaps with another action or step already selected, planned, or present in the output. Adding it would be repetitive without contributing new value. The user identifies that the same ground is covered elsewhere. \\
Partially Aligned Needs Refinement (N=26) & The action is directionally correct and in roughly the right area, but as currently framed it requires editing, narrowing, broadening, rephrasing, or better contextualization before it is useful. The user did not fully reject the action but instead modified, replaced, or adapted it to better match their specific needs. Rationales typically indicate partial alignment that needed adjustment rather than outright dismissal. \\
Tool Or System Capability Limitation (N=15) & The user recognizes that the system lacks the technical capability to properly execute the action — for example, it cannot render tables, charts, diagrams, code snippets, or complex visual layouts. The action is rejected not because of its intent but because of known constraints in the tool's output format or medium. \\
Non Substantive Test Entry (N=6) & The rationale is a placeholder, demo check, or test submission with no substantive rejection reasoning. These are system testing artifacts rather than genuine user evaluations. \\
\end{longtable}
\rowcolors{2}{}{}
\normalsize

\clearpage
\scriptsize
\centering
\captionsetup{singlelinecheck=false}
\setlength{\LTleft}{\fill}
\setlength{\LTright}{\fill}
\renewcommand{\arraystretch}{1.2}
\setlength{\tabcolsep}{6pt}
\rowcolors{2}{gray!8}{white}
\begin{longtable}{>{\raggedright\arraybackslash}p{0.27\textwidth} >{\raggedright\arraybackslash}p{0.67\textwidth}}
\caption{Uncovered reasons that actions were selected by participants based on qualitative analyses of their rationales with Claude-4.6 Opus, sorted by N descending. The descriptions are the exact ones generated by Claude. \label{appendix:table:good_rationale_cluster_full}}\\
\toprule
Cluster title & Cluster description \\
\midrule
\endfirsthead
\toprule
Cluster title & Cluster description \\
\midrule
\endhead
\bottomrule
\endfoot
Goal Aligned (N=541) & User selects the action because it directly addresses, matches, or is essential to their core question, research objective, stated goal, or primary intent. The rationale explicitly confirms the action hits the heart of what was asked, is central to the query, or fulfills the user's principal requirement. \\
Improves Clarity And Understanding (N=483) & User selects the action because it deepens comprehension or makes complex content more accessible—through definitions, step-by-step explanations, analogies, concrete examples, visualizations, illustrative case studies, causal explanations, cross-domain connections, or richer mechanistic insight—reducing cognitive barriers and enriching the user's grasp of the material. \\
Scope And Focus Control (N=379) & User selects the action because it narrows, bounds, or tightens the scope of the response—filtering out off-topic, overly broad, tangential, or irrelevant content—to keep the output precisely targeted and free of noise. Includes narrowing by domain, timeframe, specificity, or excluding distracting material. \\
General Approval (N=365) & User selects the action with minimal, generic, or tentative endorsement—brief positive assessments like 'it's fine', 'sounds reasonable', or 'could help'—without articulating a substantive underlying reason. Includes passive acceptance, low-elaboration agreement, hedged inclusion, and consistency-based re-endorsement of prior choices. \\
Actionable Implementation (N=355) & User selects the action because it provides practical, implementable guidance—concrete steps, checklists, code snippets, decision frameworks, troubleshooting tips, tool recommendations, or deployment guides—that bridge the gap between knowledge and execution, enabling the user to directly act on the information. \\
Improves Structure (N=293) & User selects the action because it improves the logical organization, coherence, narrative flow, or sectioning of the output—making it easier to navigate, follow a progression, or consume through well-structured layouts such as problem-solution framing, step-by-step ordering, or modular sections. \\
Comparison And Decision Support (N=289) & User selects the action because it enables side-by-side comparisons, highlights pros and cons, surfaces trade-offs, or provides decision frameworks and evaluative criteria that help differentiate alternatives and support informed decision-making among competing methods, tools, or approaches. \\
Improves Rigor And Evidence (N=275) & User selects the action because it strengthens the evidentiary quality, credibility, or analytical depth of the output—through empirical benchmarks, quantitative metrics, formal or mathematical rigor, validated methods, peer-reviewed sources, quality filtering, or reproducible validation—increasing confidence in claims. \\
Identifies Gaps And Future Directions (N=241) & User selects the action because it surfaces research gaps, open questions, known limitations, unresolved challenges, failure modes, or future directions—providing a forward-looking roadmap that helps the user understand what is missing, position novel contributions, or guide further investigation. \\
Concise Scannable Format (N=240) & User selects the action because it produces output that is quick to scan, digest, or reference—through bullet summaries, comparison tables, key-takeaway sections, FAQs, compact formats, structured visual layouts, or executive summaries—enabling efficient consumption and easy revisiting of key points. \\
Exploratory Curiosity (N=233) & User selects the action out of intellectual curiosity, willingness to experiment, or openness to unexpected value—driven by interest in seeing what the system produces, exploring a novel angle, testing the model's capabilities, or recognizing a proactively surfaced insight they had not considered. The selection is driven by discovery rather than confident expectation of utility. \\
Foundational Context (N=231) & User selects the action because it provides prerequisite knowledge, foundational definitions, theoretical grounding, historical context, or a baseline overview necessary to properly understand the main topic—establishing shared grounding before engaging with deeper or more advanced material. \\
Real World Grounding (N=210) & User selects the action because it connects findings to tangible real-world contexts—deployment realities, practical feasibility constraints, hardware limitations, cost considerations, case studies, institutional challenges, or concrete application scenarios—making abstract content credible and grounded in implementation reality. \\
Completeness And Coverage (N=150) & User selects the action because it broadens the scope of the output to ensure thoroughness—covering multiple dimensions, diverse perspectives, emerging topics, adjacent areas, or important aspects that would otherwise leave the response incomplete or create blind spots. \\
Expertise Calibration (N=128) & User selects the action because it calibrates depth to their existing knowledge level—skipping basics, avoiding redundant introductions, or omitting content the user already possesses. The underlying motivation is efficiency by not re-covering familiar ground and pitching content at the appropriate expertise level. \\
Personalization To Background (N=120) & User selects the action because it is specifically tailored to their personal expertise, disciplinary background, research niche, or professional profile—making the content directly relevant to their particular context and the way they naturally engage with the material. \\
Format Tone Style (N=101) & User selects the action because they endorse, prefer, or accept the proposed writing tone, stylistic register, communication style, or presentational format—valuing that the stylistic choice (e.g., assertive, academic, direct, blog-post style) fits their needs, aids persuasiveness, or does not conflict with their expectations. \\
Balanced Critical Analysis (N=70) & User selects the action because it provides a balanced, critical perspective—examining both strengths and limitations, opportunities and challenges, or benefits and risks—yielding a more nuanced, credible, and intellectually honest assessment rather than a one-sided view. \\
Ensures Recency (N=58) & User selects the action because it prioritizes recent, up-to-date, or state-of-the-art information—ensuring the output reflects the latest developments, breakthroughs, or active progress in a fast-moving field. \\
Supplementary Enrichment (N=51) & User selects the action as a non-essential but valued addition—a nice-to-have, helpful extra, or complementary extension that incrementally enriches the output without being strictly critical to the core goal. \\
Validates Known Understanding (N=15) & User selects the action because it confirms, aligns with, or validates something they already know to be true, important, or challenging—demonstrating that the model has correct domain understanding and that the action's framing or assumptions are well-suited given existing expertise. \\
Non Substantive Input (N=5) & User provided a test, demo, placeholder, or non-meaningful rationale that does not convey a genuine reason for selecting the action. These entries are system tests or artifacts rather than deliberate selection reasoning. \\
Overrides Model Suggestion (N=4) & User explicitly rejects the model's proposed actions and substitutes their own alternative. The selection reflects dissatisfaction with the model's suggestions rather than endorsement of a specific action. \\
Other (N=1) & The rationale explicitly states that the action is not relevant and does not align with the query's intent, which is a rejection of the model's suggestion. Since the provided taxonomy lacks a specific cluster for 'rejection of irrelevant suggestions' and the 'overrides\_model\_suggestion' cluster is empty/undefined for this context, 'Other' is the only accurate fit for a negative selection. \\
\end{longtable}
\rowcolors{2}{}{}
\normalsize

\clearpage
\hypersetup{
    linkcolor=white,
    citecolor=white,
    urlcolor=white
}

\lstset{
  literate={<}{{<}}1
           {>}{{>}}1
}

\begin{prompt}[title={Prompt \thetcbcounter: Generic Action Generation Prompt}, label=prompt:action_generation_generic]

\scriptsize

The user is now asking:
<query>
[query]
</query>

This query will eventually be fed into a system called PersonalizedQA that executes:\\
1. retrieval: searches for research papers\\
2. organization: outlines sections for the final response to include\\
3. generation: produces text for each of these sections\\

To help PersonalizedQA personalize responses based on the user's information, come up with a list of personalization strategies that the system should follow. Each personalization strategy should specify two requirements:\\
1. What kind of response the user will experience (Qualitative Personalization)\\
2. How the system should behave at each step (Implementation Personalization)\\

The qualitative personalization label is based on how the response will be personalized to the user at a qualitative level:
<qualitative personalization strategies>
[description of Content, Style, ...]
</qualitative personalization strategies>\\

The implementation personalization label is based on the three-step execution of ScholarQA, categorized as:
<implementation personalization strategies>
[description of Retrieval, Generation, ...]
</implementation personalization strategies>\\

<format instructions>
Generate your output as a JSON object with the single key "outputs" which has a list of personalization strategy objects, with exactly four elements total. The strategy objects should be ranked from most likely to least likely to be selected by the user, but all should have high predicted selection rates. Each strategy object should have five keys: 1) a string "strategy" as a brief high-level requirement for the final output that would lead to a more helpful response that is personalized to this user; 2) a string "tldr" with an extremely brief version of the "strategy" (no more than fifteen words); 3) a string "explanation" which explains how you decided to come up with this strategy and why this output is ranked where it is in the list; 4) a string "qualitative\_strategy" categorizing how the output will be affected qualitatively; and 5) a string "implementation\_strategy" categorizing how the instruction will be implemented in PersonalizedQA. All requirements should be a concise sentence (<30 words) in the exact form "I can... [action to take], which might help you... [predicted help]".
</format instructions>\\

<action instructions>
- If you are not confident the action is possible (e.g. if you do not know if there are papers that exist on a topic X in search\_add or search\_refine), use careful, hedged wording to avoid overclaiming, like "I can see if there are papers on X". Always hedge on search actions, but only when you are not fully confident on organization and generation.\\
- Not every strategy needs to involve adding information. You can also propose strategies so that the user can save time, like by prioritizing certain papers in search\_add, omitting background or definition sections, and not explaining basic concepts. If you add these, frame them positively. Instead of saying what you will not do, say what you will do instead. (e.g. instead of saying "I will not talk about X", say "I will focus more on Y", instead of saying "I will skip basic definitions", say "I will only define advanced terminology")\\
- The system can only generate text, so do not suggest actions that involve generating other modalities like images, audio, videos, diagrams, etc. Only suggest actions that can be executed in HTML. You can propose simple output formats like code, math, and tables since they can be rendered in HTML, but nothing too complex otherwise.
</action instructions>\\

<strategy instructions>
- Each strategy should be self-contained, meaning that it can be understood on its own by PersonalizedQA. For example, instead of saying "Since you like evaluation, I can do the same", mention what "doing the same" will entail in your response (e.g. I can add summarization to my search terms). Instead of saying you will propose an idea, disambiguate a term, or add a section, specifically mention what that idea, term, or section will be.\\
- Each strategy should give a specific action of what you could do, not hedging between multiple actions. For example, instead of saying "I can disambiguate DFS to mean depth-first search or distributed file system", pick one of these, like "I can disambiguate DFS to mean depth-first search".\\
- Each strategy needs to follow the format of "I can..., which might help you...".\\
- Each strategy should only be a single sentence\\
- Do not include citations in the text of the strategy (i.e. numbers in square brackets)\\
- Do not introduce information, jargon, or concepts that you think the user does not already know about. For example, if you think the user does not know about machine learning and asks "What are neural networks", none of your personalization strategies should include the phrase "backpropagation" if you are not fully confident the user knows what this means. Your language must be extremely simple and easy to understand.\\
- Each strategy must be extremely diverse. Do not repeat information across personalization strategies.\\
- Be extremely creative when coming up with strategies; they should be different depending on the query. For example, it is easy to always say "I will add examples" and PersonalizedQA may already do that, so propose strategies that are unlikely and uncommon for PersonalizedQA to already do.\\
- Each strategy should add extra, useful information to the response. This will likely involve inferring the intent behind the user's query. For example, if you think the user is trying to implement something, add more practical suggestions. If you think the user is brainstorming, enter a more creative mode to help them come up with ideas.\\
</strategy instructions>\\

<tldr instructions>
- Each "tldr" should be a very brief version of the strategy (around 15 words)\\
- Each "tldr" should not be in first-person. It should be a command. For example: "Include papers on X" (search\_add), "Focus the scope for X" (search\_refine), "Add section on X" (organization), "Do X in the response" (generation)\\
- Each "tldr" should be uniquely distinct from other "tldr"s.\\
- Each "tldr" should be understandable on its own without the strategy. So be very clear and specific while remaining concise.\\
- Each "tldr" should have a one-to-one mapping with the strategy, meaning it should contain all of the key information in the strategy that makes it unique. For example, if a strategy says "I will look for papers on multi-agent workflows in a scientific domain", a good "tldr" is "Look for multi-agent science papers" since it has all the salient terms. A bad "tldr" is "Look for papers involving agents" since it misses out on keywords.\\
- Each "tldr" should have no ambiguity and it should be clear as to what the system will do. Do not use terminology with multiple interpretations. For example, instead of saying "add examples", say "add empirical/analogy/real-world examples". Instead of saying "Find papers with empirical results", say "Find papers with results on benchmark datasets".\\
</tldr instructions>
\end{prompt}

\begin{prompt}[title={Prompt \thetcbcounter: Paper-Based Action Generation Prompt}, label=prompt:action_generation_paper]

\scriptsize

Here are a list of inferences about a user. The numbered inference is a high-level inference, while the sub bullet points provide evidence for these inferences:
<profile>
[profile]
</profile>\\

They are now asking:
<query>
[query]
</query>\\

...\\

<citation instructions>
- Make sure "inference\_number" is the numbered inference in <profile></profile> from which the strategy was derived.\\
- Do not hallucinate the inference number
</citation instructions>\\

<personalization instructions>
- When designing a personalization strategy, do not just consider what the researcher knows or prefers, but also what the researcher does NOT know or does NOT prefer. For example, if a cybersecurity researcher asks for papers genetic sequencing, we likely need to add more background information for this user. This should involve adding a preliminary background section in "organization" or using simple terminology in "generation". On the converse, if the user is an expert in a topic, state that you will jump straight into the main content (e.g. ignore introduction sections and not define basic terms) to help save the user time.\\
- Do not force the personalization strategies to be specific. The specificity of each strategy should depend on how similar the query is to the user's profile. For example, if a user works on knowledge graphs and the query relates to knowledge graphs, the personalization strategies should be very specific based on the user's profile, outlining more concrete actions to take. However, if this same user with interests in knowledge graphs asks about computer vision, the actions to take in the personalization strategies should be more high-level.\\
- Do not always try to directly copy the user's profile when making requirements. For example, if a user's profile says they are interested in a specific psychological construct and you want to give a strategy involving this (e.g. I will connect the explanation to Ebbinghaus's learning curve), do not mention the specific construct. You should instead write more broadly (e.g. I will connect the explanations to memory constructs).\\
- If the query is very aligned with the user's profile, provide much more concrete suggestions for personalization. But if the query is quite dissimilar, keep the personalization suggestions very high-level and broad.\\
- When personalizing strategies, do not hyper-personalize. Each strategy must still be highly related to the query. For example, if the user asks a query about the medical domain and you know they work on physics, the strategy should not be "I will cover physics". In this case, the user knows about physics but asked a question about medicine, so they want strategies that relate to both the profile and the query equally\\
- To reiterate, each strategy must be mostly related to the profile and highly related to the query.\\
- The tldr for each step should also be personalized to the user's profile
- Ensure there is an even split between where the implementation of the personalization strategy in PersonalizedQA should occur---namely an even distribution between "retrieval", "organization", and "generation" in the label "implementation\_strategy".
</personalization instructions>\\

<action instructions>
[same as above]
</action instructions>\\

<strategy instructions>
- Each strategy should be self-contained, meaning that it can be understood on its own by PersonalizedQA, as the system will not have access to the user's profile. For example, instead of saying "Since you like evaluation, I can do the same", mention what "doing the same" will entail in your response (e.g. I can add summarization to my search terms)\\
- Each strategy should give a specific action of what you could do, not hedging between multiple actions. For example, instead of saying "I can disambiguate DFS to mean depth-first search or distributed file system", pick one of these, like "I can disambiguate DFS to mean depth-first search".\\
- Each strategy needs to follow the format of "I can..., which might help you...".\\
- Each strategy must concretely explain how this strategy relates to the user's profile. It should explain why this relates to the user's profile, papers, etc.\\
- Each strategy should only be a single sentence\\
- Do not include citations in the text of the strategy (i.e. numbers in square brackets)\\
- Do not introduce information, jargon, or concepts that the user does not already know about. For example, if the user does not know about machine learning and asks "What are neural networks", none of your personalization strategies should include the phrase "backpropagation" if you are not fully confident the user knows what this means. Your language must be extremely simple and easy to understand.\\
- Each strategy must be extremely diverse. Do not repeat information across personalization strategies.\\
- Be extremely creative when coming up with strategies; they should be different depending on the query. For example, it is easy to always say "I will add examples" and PersonalizedQA may already do that, so propose strategies that are unlikely and uncommon for PersonalizedQA to already do.\\
- Each strategy should add extra, useful information to the response. This will likely involve inferring the intent behind the user's query. For example, if you think the user is trying to implement something, add more practical suggestions. If you think the user is brainstorming, enter a more creative mode to help them come up with ideas.
</strategy instructions>\\

<tldr instructions>
[same as above]
</tldr instructions>
\end{prompt}

\begin{prompt}[title={Prompt \thetcbcounter: Rule-Based Action Generation Prompt}, label=prompt:action_generation_stated]

\scriptsize

Here are a list of stated preferences from the user that describe the types of actions and types of final responses that the user who asked the query specifically liked and did not like in our system. This user came up with these preferences after having already interacted with our system for a considerable amount of time.\\
        
Each stated preference is prefixed with the phrase 'Preference [preference number]:' 
Each action history item is formatted as a 'Preference', 'Explanation' pair, where 'Preference' describes the user's preference and 'Explanation' gives the reasoning behind it:\\
<stated preferences>
[stated preferences]
</stated preferences>\\

They are now asking:
<query>
[query]
</query>\\

... [the rest is the same as the paper-based prompt] ...\\
\end{prompt}

\begin{prompt}[title={Prompt \thetcbcounter: History-Based Action Generation Prompt}, label=prompt:action_generation_revealed]

\scriptsize

Here is an action history that describes, for the same user who asked the current query, the previous queries that the user asked the system, the previous actions that were generated for each query, and whether or not the user selected or did not select the action as one for PersonalizedQA to execute. If an action was selected, it means that the user liked the action, and vice versa if it was not selected. Each action history item is formatted as a 'Query', 'Action', and 'Was Selected' triplet, where 'Was Selected = 1' if the action was selected (and = 0 otherwise):\\
<action history>
[action history]
</action history>\\

They are now asking:
<query>
[query]
</query>\\

... [the rest is the same as the paper-based prompt] ...\\
\end{prompt}

\begin{prompt}[title={Prompt \thetcbcounter: Deduplication Prompt}, label=prompt:deduplication]

\scriptsize

The user asked the query: [query]\\

To assist them, we generated two plans with steps that could provide more useful information in the answer. Each plan is in the form <Step step\_number>[step\_description]</Step step\_number>, where "step\_number" is an integer and "step\_description" is a string.\\

Here is the first plan:
<plan1>
[list of generic actions]
</plan1>\\

Here is the second plan:
<plan2>
[list of paper-based actions]
</plan2>\\

<task>
- Determine if any of the steps in plan1 and plan2 are EXACTLY semantically equivalent. Semantic equivalence means that following the instructions in the two steps would lead to the exact same answer answers\\
- For example, the steps "Include evaluation suite and dataset papers for long-form QA" and "Include only long-form QA evaluation resources" are semantically equivalent because when executed, both steps would return evaluation resources for long-form QA.\\
- For example, the steps "Limit to academic long-form QA datasets" and "Skip broad QA and generic benchmark introductions" are not semantically equivalent because when executed, the first step would focus the content on long-form QA datasets while the second step would skip an introduction.\\
- It is highly possible that the two plans will have no steps that are semantically equivalent. In this case, return an empty list.\\
</task>\\

<formatting>
- Return a JSON object with the key "duplicated\_steps" which has a list of pairwise step objects that are semantically equivalent in plan1 and plan2\\
- Each pairwise step object should have they keys "plan\_1\_step\_number" and "plan\_2\_step\_number" which have the "step\_number" of the steps in plan1 and plan2 that are equivalent. "plan\_1\_step\_number" corresponds to the step in plan1 and "plan\_2\_step\_number" corresponds to the step in plan2.\\
- return an empty JSON if there are no semantically equivalent steps.\\
</formatting>

...
\end{prompt}

\begin{prompt}[title={Prompt \thetcbcounter: Action Cluster Generation Prompt}, label=prompt:cluster_generate_action]

\scriptsize

\#\# Task: Cluster Report-Modification Actions by High-Level Strategy\\

You are analyzing the behavior of a deep research system that generates scientific reports.\\

After a user asks a research question, the system proposes a set of follow-up **actions** it could take (beyond simply answering the query) in order to improve the usefulness, clarity, rigor, or presentation of the final report. These actions represent optional modifications to the report generation process (e.g., restructuring the report, adding comparisons, refining literature scope, including technical detail, changing formatting, etc.).\\

Your task is to cluster these actions by their **high-level strategic intent** — specifically:\\

> **What kind of modification to the report this action is trying to make**\\

Cluster based on *how the action would change or improve the report*, **NOT** based on:\\
- the scientific domain\\
- the specific dataset or benchmark mentioned\\
- the topic area (e.g., medicine, robotics, climate)\\
- the literal wording of the action\\

For example:\\
- “Construct a comparison table for medical QA benchmarks”
  should be clustered under something like `comparison\_table`  
  **NOT** under anything related to medicine or QA.\\

This is because the goal is to understand patterns such as:\\
- Users respond well to comparison tables\\
- Users respond poorly to expanded literature searches\\
- Users like methodological deep-dives\\
- Users dislike structural reorganization\\
—not patterns tied to subject matter.\\

\#\#\# Clustering Requirements\\

- Create clusters that represent **distinct high-level report-modification strategies**.\\
- Clusters should be mutually exclusive in meaning wherever possible.\\
- Slight semantic overlap is acceptable, but avoid redundant clusters that represent the same strategic behavior.\\
- Prefer merging similar strategic intents into an existing cluster rather than creating a new one.\\
- Avoid creating catch-all clusters such as "other", "misc", etc.\\
- Cluster sizes may be uneven.\\
- Each action must appear in **exactly one** cluster.\\
- Do **not** omit or invent any action\_ids.\\
- Use only action\_ids that appear in the list below (range {{min\_action\_id}} to {{max\_action\_id}}).\\
- Each cluster must contain **unique** action\_ids (no duplicates across clusters).\\

\#\#\# Cluster Naming Rules\\

- `cluster\_name` must:\\
  - be in **camel\_case**\\
  - describe the strategic report modification  
    (e.g., `comparison\_table`, `add\_technical\_detail`, `refine\_search\_scope`, `restructure\_outline`, `format\_visualization`)\\
  - function like a short semantic ID\\
  - avoid domain-specific language\\

\#\#\# Output Format\\

Respond with a JSON object with a single key `"clusters"` whose value is an array of objects.\\

Each object must contain:\\
- `"cluster\_name"` (string)\\
- `"cluster\_description"` (clear explanation of the strategic report modification this cluster represents)\\
- `"action\_ids"` (array of integers)\\

\#\#\# Determinism Guidelines\\

When an action could reasonably fit in multiple clusters:\\
1. Assign it based on the **primary report-modification strategy**.\\
2. Prefer an existing cluster over creating a new one.\\
3. Create a new cluster only if the action represents a genuinely new type of strategic modification.\\

\#\#\# List of actions (id in brackets, one per line)\\

[insert actions]
\end{prompt}

\begin{prompt}[title={Prompt \thetcbcounter: Rationale Cluster Generation Prompt}, label=prompt:cluster_generate_rationale]

\scriptsize
\#\# Task: Cluster User Rationales for REJECTED Actions (action\_score=0)\\

You are analyzing a deep research system where users are shown proposed follow-up actions to improve a scientific report. For each action, users can **select** it (action\_score=1) or **reject** it (action\_score=0). When they reject an action, they provide a **rationale** (action\_cot) explaining why they did *not* choose it.\\

Your task is to cluster these **user rationales** (the reasons users give for *rejecting* an action) by their **high-level theme or pattern** — specifically:\\

> **What kind of reasoning or objection does this rationale represent?**\\

Cluster based on the *type or pattern of user reasoning*, **NOT** based on:\\
- the scientific domain or topic\\
- the specific action text\\
- the literal wording\\
- the user’s specific query, project, or constraints (make clusters query-agnostic)\\
- the surface form of the concern (e.g., “engineering”, “deployment”, “implementation”, “data collection”, “compliance”)\\

Your clusters are meant to diagnose **why suggested actions are not preferred** and what to improve about the action generator (relevance, focus, scope, cost, clarity, etc.). Therefore, you must **abstract** each rationale into an underlying rejection reason that would generalize across many contexts.\\

For example:\\
- "no need for this" and "No need for that"
  might cluster under something like `unnecessary\_for\_goal`\\
- "Do not need beyond metrics for this" and "Just benchmark is what I need"
  might cluster under something like `scope\_too\_broad`\\
- "This goes into deployment details" and "I don't want engineering implementation steps"\\
  should **not** become `avoid\_deployment` / `avoid\_engineering`; instead they might map to `distracts\_focus` or `wrong\_level\_of\_abstraction` depending on intent.\\
- "This would take too long / too much work"
  might map to `too\_costly\_for\_value` (not `needs\_more\_time\_for\_experiments`).\\
- "I don't want debugging/testing/implementation details"
  should **not** become `avoid\_debugging` / `avoid\_testing` / `avoid\_implementation`; these almost always map to a broader underlying issue like `wrong\_level\_of\_abstraction`, `distracts\_focus`, or `intent\_misaligned`.\\

The goal is to surface patterns such as:
- Users reject because the action is **misaligned with their current goal** (`unnecessary\_for\_goal`, `wrong\_direction`)\\
- Users reject because it **distracts focus / wrong level of abstraction** (`distracts\_focus`, `too\_implementation\_heavy`, `too\_academic\_for\_need`)\\
- Users reject because it is **too broad / too deep / too much** (`scope\_too\_broad`, `too\_many\_steps`, `would\_bloat\_report`)
- Users reject because it is **too costly for the expected value** (`too\_costly\_for\_value`, `low\_roi`)\\
- Users reject because it is **unclear or underspecified** (`unclear\_request`, `missing\_actionable\_details`)\\

\#\#\# Clustering Requirements\\

- Create clusters that represent **distinct, query-agnostic, high-level rejection patterns**.\\
- Clusters should be mutually exclusive in meaning wherever possible.\\
- Slight semantic overlap is acceptable, but avoid redundant clusters.\\
- Prefer merging similar rationale types into an existing cluster.\\
- Avoid catch-all clusters such as "other", "misc", etc.\\
- Cluster sizes may be uneven.\\
- Prefer **fewer, broader clusters** over many narrow ones. If you find yourself creating many clusters that differ only by *topic* (e.g., `avoid\_X`), you are doing it wrong—merge them by the underlying rejection reason.\\
- Each rationale must appear in **exactly one** cluster.\\
- Do **not** omit or invent any rationale\_ids.\\
- Use only rationale\_ids that appear in the list below (integers from 0 to {{max\_rationale\_id}}).\\
- Each cluster must contain **unique** rationale\_ids (no duplicates across clusters).\\

\#\#\#\# Abstraction rule (critical)\\
- Before assigning a rationale to a cluster, rewrite it mentally into: **“Underlying reason for rejection is X.”**\\
- If a rationale mentions a *specific* concern-area (engineering, deployment, infra, citations, experiments, metrics, etc.), map it to the **underlying effect on the user’s goal** (distracts focus, wrong depth, too costly, not needed, unclear, redundant), not the concern-area itself.\\
- Do not create clusters that are essentially “user doesn’t want to focus on <topic>”. Those should be merged into broader patterns like `distracts\_focus`, `wrong\_level\_of\_abstraction`, `scope\_mismatch`, or `not\_relevant\_now`.\\
- In particular, **do not create** separate clusters like `avoid\_debugging`, `avoid\_implementation`, `avoid\_deployment`, `avoid\_engineering`, `avoid\_testing`. Those are usually the same underlying rejection pattern: the action is **misaligned with the user’s intent** and/or at the **wrong level of abstraction**.\\

...[same name rules and output format as above]...\\

\#\#\# List of rationales (id in brackets, one per line)

[insert rationales]

\end{prompt}

\begin{prompt}[title={Prompt \thetcbcounter: Cluster Merging Prompt}, label=prompt:cluster_merge]

\scriptsize

\#\# Task: Merge Per-User Action Clusters into a Global Taxonomy\\

You are given a set of **per-user clusters** from a deep-research action clustering step. Each user had their actions clustered independently, so cluster names and descriptions may overlap or differ slightly across users (e.g., one user might have "comparison\_table" and another "addComparisonTable" for the same concept).\\

Your task is to **merge** these into a single global taxonomy of clusters:\\

1. **Group** user clusters that represent the **same high-level report-modification strategy** into one merged cluster.\\
2. **Assign** each user cluster (by its id) to exactly one merged cluster.\\
3. **Name** each merged cluster in camel\_case, with a clear global `cluster\_name` and `cluster\_description` that captures the unified strategy.\\
4. Every user cluster id must appear in exactly one merged cluster's `cited\_user\_cluster\_ids` list.\\

\#\#\# Input: Per-User Clusters (id → name, description)\\

Each line describes one cluster from one user. Format:
`[id] cluster\_name — cluster\_description`\\

[insert clusters]\\

\#\#\# Output Format\\

Respond with a JSON object with a single key `"merged\_clusters"` whose value is an array of objects.\\

Each object must contain:\\
- `"cluster\_name"` (string, camel\_case): global name for this merged cluster\\
- `"cluster\_description"` (string): clear explanation of the report-modification strategy\\
- `"cited\_user\_cluster\_ids"` (array of strings): list of ids from the input that belong to this merged cluster (e.g. `["u0\_c0", "u1\_c2", "u2\_c0"]`)\\

\#\#\# Guidelines\\

- Merge user clusters that are semantically the same strategy even if names/descriptions differ.\\
- Prefer a single merged cluster over fragmenting into many tiny clusters.\\
- Do not invent or omit any id; every id from the input must appear in exactly one `cited\_user\_cluster\_ids` list.\\
- Keep merged cluster names and descriptions domain-agnostic and strategy-focused (same style as the original clustering task).\\

\end{prompt}

\begin{prompt}[title={Prompt \thetcbcounter: Cluster Labeling Prompt}, label=prompt:cluster_label]

\scriptsize

You are a precise classifier. Classify the action into exactly one cluster from the provided cluster catalog. Use semantic meaning, not keyword overlap. Return only JSON with keys: cluster\_title, explanation.\\

[insert clusters and action]\\
\end{prompt}

\begin{prompt}[title={Prompt \thetcbcounter: Action Prediction Prompt}, label=prompt:action_prediction]

\scriptsize

<task>
You are an expert at evaluating which actions a user would want a Deep Research system to take.\\

<context>
A user asked a Deep Research system called MyPaperQA a question related to research in computer science, eventually wanting to see a multi-section report that searches for and synthesizes literature from scholarly search engines to answer the question. Before directly answering the question and writing the report, MyPaperQA generates a list of actions: extra steps the system could take while answering the question to make the report more useful and tailor to the user's individual needs (e.g. For the question "Give me an overview of question answering benchmarks", an action could be "Focus on medical question answering datasets"). The user saw the list of these actions and selected which ones they did want and did not want MyPaperQA to execute when answering the query.
</content>\\

Given a query and one of the actions that MyPaperQA proposed to execute while answering that query, your job is to classify whether the user would want the system to execute that action. To figure this out, you will analyze the user's previous judgments of query, action pairs. You will uncover patterns, trends, and behaviors in the past actions that the user liked and disliked, then try to apply the most relevant trends to the current query and action.
</task>\\

Here are queries and actions this same user previously asked MyPaperQA in chronological order, along with whether or not they selected this action as one of the steps for MyPaperQA to execute for the query:\\
<history>
[insert action selection history]
</history>\\

Here is the current query asked by the user:\\
<query>
[insert query]
</query>\\

Here is the current action for the query that MyPaperQA generated and was shown to the user:\\
<action>
[insert action]
</action>\\

<format>
Generate a JSON with two keys: 1) "was\_selected" - a 0/1 label with your prediction for whether or not the user selected this action as one for the system to execute (0 means not selected, 1 means selected); and 2) "explanation" - an explanation for your decision. The format is as follows: [insert JSON format]
</format>

\end{prompt}

\begin{prompt}[title={Prompt \thetcbcounter: Action Prediction Ranking Prompt}, label=prompt:action_multi_classification]

\scriptsize

<task>
You are an expert at evaluating which actions a user would want a Deep Research system to take.\\

<context>
A user asked a Deep Research system called MyPaperQA a question related to research in computer science, eventually wanting to see a multi-section report that searches for and synthesizes literature from scholarly search engines to answer the question. Before directly answering the question and writing the report, MyPaperQA generates a list of actions: extra steps the system could take while answering the question to make the report more useful and tailor to the user's individual needs (e.g. For the question "Give me an overview of question answering benchmarks", an action could be "Focus on medical question answering datasets"). The user saw the list of these actions and selected which ones they did want and did not want MyPaperQA to execute when answering the query.
</content>\\

Given a query and a list of actions that MyPaperQA proposed to execute while answering that query, your job is to classify whether the user would want the system to execute each action. To figure this out, you will analyze the user's previous judgments of query, action pairs. You will uncover patterns, trends, and behaviors in the past actions that the user liked and disliked, then try to apply the most relevant trends to the current query and actions.
</task>\\

Here are queries and actions this same user previously asked MyPaperQA in chronological order, along with whether or not they selected this action as one of the steps for MyPaperQA to execute for the query:
<history>
[insert action selection history]
</history>\\

Here is the current query asked by the user:
<query>
[insert query]
</query>\\

Here is the list of actions for the query that MyPaperQA generated and was shown to the user:
<actions>
[insert actions]
</actions>\\

<format>
Generate a JSON with one key: 1) "predictions" - a list of predictions for each of the above actions. Each element of "predictions" should be a JSON with two keys: 1) "was\_selected": a 0/1 label with your prediction for whether or not the user selected this action as one for the system to execute (0 means not selected, 1 means selected); and 2) "explanation" - an explanation for your decision. The format is as follows: [insert JSON]\\
Do not generate anything else
</format>

\end{prompt}

\begin{prompt}[title={Prompt \thetcbcounter: Action Prediction Pairwise Prompt}, label=prompt:action_ranking]

\scriptsize

<task>
You are an expert at evaluating which actions a user would want a Deep Research system to take.\\

<context>
A user asked a Deep Research system called MyPaperQA a question related to research in computer science, eventually wanting to see a multi-section report that searches for and synthesizes literature from scholarly search engines to answer the question. Before directly answering the question and writing the report, MyPaperQA generates a list of actions: extra steps the system could take while answering the question to make the report more useful and tailor to the user's individual needs (e.g. For the question "Give me an overview of question answering benchmarks", an action could be "Focus on medical question answering datasets"). The user saw the list of these actions and selected which ones they did want and did not want MyPaperQA to execute when answering the query.
</content>\\

Given a query and a list of actions that MyPaperQA proposed to execute while answering that query, your job is to classify whether the user would want the system to execute each action. To figure this out, you will analyze the user's previous judgments of query, action pairs. You will uncover patterns, trends, and behaviors in the past actions that the user liked and disliked, then try to apply the most relevant trends to the current query and actions.
</task>\\

Here are queries and actions this same user previously asked MyPaperQA in chronological order, along with whether or not they selected this action as one of the steps for MyPaperQA to execute for the query:
<history>
[insert action selection history]
</history>\\

Here is the current query asked by the user:
<query>
[insert query]
</query>\\

Here is the first list of actions (action list A) for the query that MyPaperQA generated and was shown to the user:
<action list A>
[insert action A]
</action list A>\\

Here is the second list of actions (action list B) for the query that MyPaperQA generated and was shown to the user:
<action list B>
[insert action B]
</action list B>\\

<format>
Generate a JSON with two keys: 1) "selected\_list" - a string label of either "A" or "B" with your prediction for which list action the user selected for the system to execute ("A" means they selected list A, "B" means they selected list "B"); and 2) "explanation" - an explanation for your decision. The format is as follows: [insert JSON]
Do not generate anything else
</format>

\end{prompt}

\begin{prompt}[title={Prompt \thetcbcounter: Action Execution Prompt}, label=prompt:action_execution]
\scriptsize

<task>
You are an expert at evaluating which actions a user would want a Deep Research system to take.\\

<context>
A user asked a Deep Research system called MyPaperQA a question related to research in computer science, eventually wanting to see a multi-section report that searches for and synthesizes literature from scholarly search engines to answer the question. Before directly answering the question and writing the report, MyPaperQA generates a list of actions: extra steps the system could take while answering the question to make the report more useful and tailor to the user's individual needs (e.g. For the question "Give me an overview of question answering benchmarks", an action could be "Focus on medical question answering datasets"). The user saw the list of these actions and selected which ones they did want and did not want MyPaperQA to execute when answering the query. MyPaperQA then generated the report while answering the question and attempting to follow these actions. The user finally labeled whether MyPaperQA was successful or unsuccessful in executing each action.
</content>\\

Given a query, one of the actions the user told MyPaperQA to execute while answering that query, and the report generated for the query and action, your job is to classify whether the system successfully executed the action.
</task>\\

Here is the current query asked by the user:
<query>
[insert query]
</query>

Here is the current action for the query that MyPaperQA generated and was shown to the user:
<action>
[insert action]]
</action>\\

Here is the report MyPaperQA generated:
<report>
[insert report]
</report>\\

<format>
Generate a JSON with two keys: 1) "was\_successful" - a 0/1 label with your prediction for whether or not the system was successful in executing the action (0 means not successful, 1 means successful); and 2) "explanation" - an explanation for your decision. The format is as follows: [insert JSON]\\
Do not generate anything else
</format>
\end{prompt}

\begin{prompt}[title={Prompt \thetcbcounter: Rule Inference Prompt}, label=prompt:rule_inference]
\scriptsize

<task>
You are an expert at inferring rules from a history of query-action pairs.\\

<context>
A user asked a Deep Research system called MyPaperQA a question related to research in computer science, eventually wanting to see a multi-section report that searches for and synthesizes literature from scholarly search engines to answer the question. Before directly answering the question and writing the report, MyPaperQA generates a list of actions: extra steps the system could take while answering the question to make the report more useful and tailor to the user's individual needs (e.g. For the question "Give me an overview of question answering benchmarks", an action could be "Focus on medical question answering datasets"). The user saw the list of these actions and selected which ones they did want and did not want MyPaperQA to execute when answering the query.
</context>\\

Given the user's history of prior query-action pairs and whether or not the user selected the action for the query and a rationale for that decision, your job is to infer a set of rules that best describe the user's preferences on actions for MyPaperQA to execute when answering queries, and explain why these are accurate and comprehensive rules for the user, citing their prior action selections and disselections as evidence. Your eventual goal will be to use these rules to determine whether the user selected or did not select new actions for MyPaperQA to execute when answering new queries, so it is important to infer a set of informative rules that could help you make those decisions
</task>\\

Here are queries and actions this same user previously asked MyPaperQA in chronological order, along with whether or not the user selected the action for the query and a rationale for that decision:
<history>
[insert action selection history]
</history>\\

<rule instructions>
- Each rule should generally be in the format "You prefer... [insert description of preference]"\\
- Rules can be dependent on just the action or the query and the action. For example, a rule could be "You prefer bullet-point lists" if you believe that applies generally, or a rule could be "You prefer bullet-point lists for queries on evaluation" if you believe that applies specifically to queries on evaluation. You should determine whether a rule applies generally or specifically to queries, but prioritize general rules versus specific rules.\\
- Rules should not contradict each other or be redundant. For example, if a rule is "You prefer bullet-point lists", then a rule like "You prefer bullet-point lists for queries on evaluation" is redundant and should not be included. Similarly, if a rule is "You prefer bullet-point lists", then a rule like "You do not prefer long bullet-point lists" or "You prefer numbered lists over bullet-point lists" is contradictory, so only one of these rules can be returned.\\
- Rules should be diverse from each other. Do not repeat information across rules\\
- Rules should be in sentence format and easy to understand\\
- Rules should be no more than 30 words.\\
- Generate a set of rules that has maximum coverage of the user's preferences. Using these rules, we should be able to reconstruct the user's preferences as comprehensively and accurately as possible.
</rule instructions>\\

<explanation instructions>
- You will be asked to generate an explanation for each rule that you generate. Make sure you always do this\\
- Each of the explanations should cite specific examples from the user's history that support the rule. You should also point out if there are any actions in the user's history that contradict the rule if they exist. But in general, this should rarely happen and you should infer rules that do not contradict the user's preferences\\
- You can cite actions that the user did and did not select to justify each rule\\
- Each example is displayed in the form <example EXAMPLE\_INDEX>Query: ... Action: ... Was Selected: ... Rationale: ...</example EXAMPLE\_INDEX>. You should cite examples by referencing the EXAMPLE\_INDEX. For example, if the rule is "You prefer bullet-point lists", the explanation could read "Examples 1, 3, and 17 showed the user preferred bullet-point lists, while example 29 showed the user did not like numbered lists. Thus, I that this user has a preference for bullet-point lists."\\
- Each explanation should be three sentences maximum.\\
</explanation instructions>\\

<format>
Generate a JSON with two keys: 1) "rules" - a list of 5 rules as sentence-long strings that best describe the user's preferences on actions for MyPaperQA to execute when answering queries; and 2) "explanations" - a list of explanations for your decision for each rule. The format is as follows: [insert JSON]\\
Do not generate anything else
</format>

\end{prompt}

\end{document}